\documentclass[a4paper]{article}
\usepackage{adjustbox}
\usepackage[top=1.in, bottom=1.in, left=1.in, right=1.in]{geometry}
\usepackage{tabls}
\usepackage{cites}
\usepackage{epsf}
\usepackage{appendix}
\usepackage{ragged2e}
\usepackage{mathtools}
\usepackage{mdframed}
\usepackage{caption}
\usepackage{physics}
\usepackage{enumitem}
\usepackage{mathtools}
\usepackage{braket}
\usepackage{booktabs}
\usepackage{tabularx}
\usepackage{makecell}
\usepackage[flushleft]{threeparttable}
\usepackage{algorithm}
\usepackage{algpseudocode}
\usepackage{nomencl} 
\makenomenclature
\usepackage{tikz}
\usepackage{graphicx}
\usepackage{float}
\floatstyle{plaintop}
\restylefloat{table}

\floatstyle{plain}
\restylefloat{figure}
\usepackage{amssymb}
\usepackage{pifont}
\usepackage{amsmath} 

\definecolor{light-gray}{gray}{0.95}
\newcommand{\code}[1]{\colorbox{light-gray}{\small\color{blue}\textbf{\texttt{#1}}}}

\def\bmheadfont{\reset@font\fontfamily{\rmdefault}\fontsize{10bp}{12bp}\bfseries\selectfont\raggedright\boldmath}%
\newcommand\bmhead{\@startsection{subparagraph}{5}{\z@}%
                                 {6pt \@plus1ex \@minus .2ex}%
                                 {-1em}%
                                 {\bmheadfont}}

\usepackage{url}
\providecommand{\keywords}[1]
{
  \small	
  \textbf{\textit{Keywords---}} #1
}

\title{Surpassing legacy approaches to PWR core reload optimization with single-objective Reinforcement learning}
\author{
  \textbf{Paul Seurin$^1$\footnote{Corresponding author. \textit{E-mail address}: paseurin@mit.edu (P. Seurin)}, Koroush Shirvan$^1$} \\
  \\
  $^1$Massachusetts Institute of Technology \\
77 Massachusetts Avenue, Cambridge MA, 02139
\\ 
  \url{paseurin@mit.edu}, \url{kshirvan@mit.edu}
}
\begin{document}

\maketitle
\begin{abstract}

 Optimizing the fuel cycle cost through the optimization of nuclear reactor core loading patterns involves multiple objectives and constraints, leading to a vast number of candidate solutions that cannot be explicitly solved. To advance the state-of-the-art in core reload patterns, we have developed methods based on Deep Reinforcement Learning (DRL) for both single- and multi-objective optimization. Our previous research has laid the groundwork for these approaches and demonstrated their ability to discover high-quality patterns within a reasonable time frame. On the other hand, stochastic optimization (SO) approaches are commonly used in the literature, but there is no rigorous explanation that shows which approach is better in which scenario. In this paper, we demonstrate the advantage of our RL-based approach, specifically using Proximal Policy Optimization (PPO), against the most commonly used SO-based methods: Genetic Algorithm (GA), Parallel Simulated Annealing (PSA) with mixing of states, and Tabu Search (TS), as well as an ensemble-based method, Prioritized Replay Evolutionary and Swarm Algorithm (PESA). We found that the LP scenarios derived in this paper are amenable to a global search to identify promising research directions rapidly, but then need to transition into a local search to exploit these directions efficiently and prevent getting stuck in local optima. PPO adapts its search capability via a policy with learnable weights, allowing it to function as both a global and local search method. Subsequently, we compared all algorithms against PPO in long runs, which exacerbated the differences seen in the shorter cases. Overall, the work demonstrates the statistical superiority of PPO compared to the other considered algorithms.
\end{abstract}

\begin{mdframed}
\begin{table}[H]   
\nomenclature{RL}{Reinforcement Learning}
\nomenclature{PPO}{Proximal Policy Optimization}
\nomenclature{MDP}{Markov Decision Process}
\nomenclature{CO}{Combinatorial Optimization}
\nomenclature{MOO}{Multi-Objective Optimization}
\nomenclature{FT}{Friedman Test}
\nomenclature{NT}{Nemenyi Test}
\nomenclature{SO}{Stochastic Optimization}
\nomenclature{GA}{Genetic Algorithm}
\nomenclature{ES}{Evolutionary Strategy}
\nomenclature{TS}{Tabu Search}
\nomenclature{PSA}{Parallel Simulated Annealing}
\nomenclature{PESA}{Prioritized replay Evolutionary and Swarm Algorithm}
\nomenclature{PEARL}{Pareto Envelope Augmented with Reinforcement Learning}
\nomenclature{FOMs}{Figure Of Merits}
\nomenclature{FA}{Fuel Assembly}
\nomenclature{LWR}{Light Water Reactor}
\nomenclature{$L_C$}{Cycle Length}
\nomenclature{LP}{Loading Pattern}
\nomenclature{$F_{\Delta h}$}{Rod integrated peaking factor}
\nomenclature{$F_q$}{Peak pin power}
\nomenclature{$C_b$}{Boron concentration}
\nomenclature{$Bu_{max}$}{Peak pin burnup}
\nomenclature{LCOE}{Levelized Cost of Electricity}
\nomenclature{BP}{Burnable Poison}
\nomenclature{IFBA}{Integral Fuel Burnable Absorber}
\nomenclature{WABA}{Wet Annular Burnable Absorber}

\printnomenclature
\end{table}
\end{mdframed}
\section{Introduction}
\label{sec:introduction}

The Pressurized Water Reactor (PWR) Loading Pattern (LP) optimization problem has been a major area of research in the field of nuclear engineering \cite{li2022areview}. Today, one of the motivations to continue to improve LP optimization is driven by keeping existing nuclear assets competitive on the grid to continue to provide dispatchable clean energy along side renewable generation. Optimizing the fuel cycle cost via the optimization of the core LP is one way that more efficient core designs can support PWR economics. Nevertheless, the large search space as well as the numerous amount of operational and safety constraints, which on top of that vary from one reactor to the next \cite{kropaczek2019largescale}, makes this problem extremely complex. To help find high-quality solutions to this problem, a reactor physic code (e.g. SIMULATE3 \cite{rempe1989simulate}) coupled with an optimization algorithm (or a combination of several \cite{lin2014themaxmin}) adopted from the computational industry is utilized. In particular heuristic-based or Stochastic Optimization (SO) algorithms thanks to their simplicity to understand and implement, their ability to traverse the search space efficiently, and their application to problems with similar structures outside the nuclear field are leveraged \cite{li2022areview}. The most widely utilized in literature are Simulated Annealing (SA), Genetic Algorithm (GA), and Tabu Search (TS) \cite{kropaczek2019largescale}. As a result some companies tried to adopt these methodologies. For instance, Duke Energy and Southern Nuclear are sometimes leveraging FORMOSA-P \cite{kropaczek1991incore} and ROSA \cite{verhagen1997rosa}, respectively, both based on SA. However, some utilities preferred hand-designed optimization such as Constellation Energy.

Recent interest in Deep Reinforcement Learning (RL) has surged as a solution to the PWR LP optimization problem \cite{seurin2022pwr,seurin2024assessment,seurin2024multiobjective}, stemming from the inherent limitations of heuristic-based methods, such as slow convergence and convergence to suboptimal local solutions \cite{li2022areview,seurin2022pwr,seurin2024assessment}. Traditionally, heuristic-based methods guide the search using rule-of-thumb principles to accept or reject new solutions. In contrast, RL involves agents learning to follow an optimal set of decision rules through trial and error during episodes \cite{seurin2024assessment}. For example, a set of decision rules leads to the construction of a core pattern \cite{seurin2022pwr,seurin2024assessment,seurin2024multiobjective}. These decisions are computed by sampling from a parameterized policy network, with learnable weights updated during optimization. We termed our RL-based approach an "automatic supervised setting" in \cite{seurin2024multiobjective} due to its similarity to supervised learning, albeit with a policy generating samples instead of a human, while a value function evaluates their quality, guiding the generation of subsequent samples. 

Our work in \cite{seurin2022pwr,seurin2024assessment,seurin2024multiobjective} represents, to our knowledge, the first attempts to leverage RL to tackle the aforementioned problem, with \cite{seurin2024multiobjective} extending it to constrained multi-objective settings. In the former, we utilized SIMULATE3 as the reactor physics code to evaluate generated cores and conducted analyses to understand the influence of various hyperparameters on optimizing the economy (i.e., the Levelized Cost of Electricity (LCOE)) of a Westinghouse-type 4-loop PWR with seven safety and operational constraints. There, we demonstrated RL's potential to solve this problem efficiently within a short timeframe (24 hours). In \cite{seurin2024multiobjective}, we introduced Pareto Envelope Augmented with Reinforcement Learning (PEARL), a novel RL-based approach for solving Multi-Objective Optimization (MOO) problems, applied to both classical benchmarks and the aforementioned problem in unconstrained and constrained scenarios. 

To our knowledge, there are no recent rigorous explanations as to which methods is better to solve the LP problem. A former study \cite{Yamamoto1997AQC} attempted to compare Simulated Annealing (SA), Direct Search (DS), Binary Exchange (BE, i.e., performing all possible binary exchanges of an LP solution and taking the best), Genetic Algorithm (GA), and combinations of them. The author demonstrated the superiority of combining GA and BE, revealing that the global search capacity of GA combined with the local search efficiency of BE outperformed more local methods like DS, SA, and DS + BE. However, the problem addressed was very simplistic: it involved one objective ($L_{cy}$) and one constraint (the radial peaking factor $F_{x,y}$), applied to a Westinghouse type 900 MWe 3-loop PWR loaded with 157 fuel assemblies (FAs) with octant symmetry. Additionally, the study (published in 1997) leveraged older versions of SA and GA. Our work presents more complex problem scenarios (see Section \ref{sec:testsets}), each more intricate than what was studied previously. Furthermore, we included more advanced versions of PSA, TS, and ES, and introduced a deep learning-based approach with RL in the comparison. Such an approach was prohibitive with the computational power available at that time. Another work presented at a conference \cite{delipei2023reactorcore} applied PPO (and Deep Q-network (DQN) and Soft Actor Critic (SAC)) on a first-cycle core LP optimization problem of a PWR in octant symmetry. The objective was to maximize the cycle length $L_C$, while satisfying constraints relative to boron concentration $C_b$, the peak pin peaking factor $F_q$, and the rod-integrated peaking factor $F_{\Delta h}$. There, they compared it with SA and GA with the desire to integrate RL in their Modular Optimization Framework (MOF). This study compared RL and SO on a PWR core design, and the authors left the tuning and another type of Markov Decision Process (MDP) (i.e., what they called a \textit{gamification}) for future research. While PPO outperformed GA, it was still worse than SA, which was itself tuned from past research in MOF. Moreover, in this comparison, one episode terminates when a LP is complete, which is when agent filled the reactor. We argued that the best MDP is to fill a LP all at once \cite{seurin2024assessment}. Therefore, there is still a need to compare our optimal approach to SO-based heuristics but most importantly understand what are the characteristics of RL that might make it superior to them. 

In \cite{seurin2024multiobjective}, we compared PEARL with NSGA-II \cite{deb2002afast}, NSGA-III \cite{jain2014anevolutionaryI,jain2014anevolutionaryII} , multi-objective TS \cite{jaeggi2005amultiobjective} , and multi-objective SA \cite{park2009multiobjective} and empirically demonstrated its superiority for our problems out of one experiment each. We hypothesize that the capacity of PPO's policy to adapt its weight over during optimization, while the classical approaches were following rules to navigate the search space could explain its superiority. Here however, we performed a series of tests in the single-objective setting to first confirm this hypothesis and rigorously explain why RL, in particular PPO, is superior. We compare PPO against commonly used Parallel Simulated Annealing (PSA) with mixing of states (arguably the state-of-the-art approach for core loading pattern optimization  \cite{kropaczek2009copernicus,ottinger2015bwropt}), GA (more specifically the Evolutionary Strategy (ES) \cite{beyer2002evolution} variant), and a novel parallel TS on the problem defined in \cite{seurin2024assessment}. Preliminary results for the single-objective comparisons were presented at the American Nuclear Society (ANS) conference in Knoxville, TN, in April 2023 \cite{seurin2023can}. However, this paper extends the applications and comparisons with legacy approaches to new design scenarios involving a reduced batch size for feed assemblies. The contributions of this work are summarized as follows:
\begin{enumerate}
    \item 
    Proposing a novel comparison between state-of-the-art GA, PSA, TS, and an ensemble method (i.e., multiple heuristics together) PESA and RL on single-objective PWR optimization problems.
    \item 
    Stressing the behavioral differences between the legacy approaches and PPO which adapt its weights depending on the curvature of the objective space to solve a large array of problems: Getting us a step closer to solve black-box optimization efficiently.
\end{enumerate}

Sections \ref{sec:reinforcementlearningdefinitionand} and \ref{sec:stochasticoptimization} briefly introduced the algorithms utilized namely RL, and the SO-based, respectively. The section \ref{sec:inputdecision} outlines the format of the inputs utilized following the work described in \cite{seurin2022pwr,seurin2024assessment}. Section \ref{sec:testsets} briefly describe the test problems, objectives, and constraints studied here. The Section \ref{sec:resultsandanalysis} showcases the results found, following an hyper-parameter studies for SA and GA in Appendix \ref{appendix:hyperparameterforpsaandes}. Finally, the conclusions and area of future research are given in Section \ref{sec:concludingremarks}.

\section{RL and SO algorithms}
\label{sec:reinforcementlearningandstochastic}

\subsection{RL definition}
\label{sec:reinforcementlearningdefinitionand}

 Deep RL is the sub-domain of Deep Learning (DL) concerned with decision-making. There, one or several agents are learning to take optimal decisions (or actions) with the help of a reward signal produced by an emulator for games or an objective function in case of optimization. The data for training are generated on-the-fly as the agent is learning, which contrasts with supervised and unsupervised learning in which data must be carefully prepared in advance. The underlying mathematical formulation of the RL mechanisms is given in \cite{seurin2022pwr,seurin2024assessment}. There are two types of algorithms to solve a RL problem: \textit{value iteration} and \textit{policy iteration}. \textit{Policy iteration} is based on optimizing a policy (or sequence of decisions) directly by a parameterization $\pi(\cdot,\cdot |\theta)$ and solving $\max_{\pi(\cdot | \theta) \in \Pi} V(\pi(\cdot|s,;\theta))$, where $\Pi$ is the manifold of smoothly parametrized policies and the value function $V(\pi(\cdot|s;\theta))$ is the expected sum of rewards following the policy $\pi(\cdot|s;\theta) = \pi(\theta)$ starting from state $s$ (i.e., the quality of being in this state). In original policy-based methods, the parameter $\theta$ is updated by policy gradient ascent with the relation
\begin{equation}
    \theta_{k+1} = \theta_k + \eta_k \nabla_{\theta}V(\pi(\theta))
    \label{eq:policygradientupdate}
\end{equation}
where $\nabla_{\theta}V(\pi(\theta)) = \mathbb{E}(R(\tau) \Sigma_{t = 1}^{\tau} \nabla_{\theta} \log \pi_{\theta}(a_t|s_t)))$. The update in equation \ref{eq:policygradientupdate} translates the fact that direction with high expected return $R(\tau) = \sum_{t=0}^{\tau -1 }\gamma^tr(s_t,a_t)$ will be pushed with greater magnitude, where $\gamma \in [0,1]$ is a hyper-parameter added to favor short-term reward if closer to 0, $r(s_t,a_t)$ is the reward of taking action $a_t$ being in state $s_t$, and $\tau$ is the length of an \textit{episode} (i.e., the period of time in which data are collected). The estimate of the gradient can be obtained by Monte-Carlo sampling over multiple trajectories. This set of methods have the advantage of simplicity, with fewer hyperparameters to tune and scale well for parallel computing. A subclass of the latter, specifically Proximal Policy Optimization (PPO) was used in \cite{seurin2022pwr,seurin2024assessment,seurin2024multiobjective}. The goal of PPO is to update the learnable parameter $\theta$ at each step k + 1 such that:
   \begin{equation} \theta_{k+1} \in arg\max_{\theta} \frac{1}{\text{\code{ncores}} \times \code{n\_steps}}\sum_{i = 1}^{\text{\code{ncores}}}\sum_{\tau = 0}^{\code{n\_steps} - 1}( \min(\frac{\pi_{\theta}(a_{\tau}|s_{\tau}))}{\pi_{\theta_{k}}(a_{\tau}|s_{\tau})}A^{\pi_{\theta_k}}(a_{\tau},s_{\tau}),g^{*}(\epsilon,A^{\pi_{\theta_k}}(a_{\tau},s_{\tau}))\end{equation} \\
    
   \begin{equation*}- \code{VF\_coef} (V^{\pi_{\theta}}(s_{\tau}) - R_i(\tau))^2) + \text{\code{ent\_coef}} \times H(\theta) 
   \label{eq:ppoloss}
\end{equation*}
where $g^{*}(\text{\code{$\epsilon$}},A) = \{ \begin{array}{l}
    (1 + \text{\code{$\epsilon$}})A \quad \text{if} \quad A \ge 0\\
    (1 -\text{\code{$\epsilon$}})A \quad \text{if} \quad A < 0 
  \end{array} $, \code{ncores} is the number of parallel agents during optimization (i.e., number of processors assign to the optimization), \code{n\_steps} is the number of actions taken by each agent before updating the PPO loss and updating the learnable parameter $\theta$, $A^{\pi_{\theta_k}}(a_t,s_t)$ is the advantage function for policy $\pi_{\theta_k}$ taking action $a_t$ being in state $s_t$, and $H(\theta)$ is the entropy of the policy $\pi_{\theta_k}$. Each term is defined in \cite{seurin2024assessment}. and we omit to redefine them for brevity. Overall, the equation \ref{eq:ppoloss} is comprised of a policy loss, a value loss, and an entropy term associated with the hyper-parameters \code{$\epsilon$}, \code{VF\_coef}, and \code{ent\_coef}, respectively. See also \cite{seurin2024assessment} for a thorough description of the hyper-parameters.

 While interacting with a reactor physic code (there SIMULATE3 \cite{rempe1989simulate}), which served as their acting environment, RL agents were trained to build a LP in \cite{seurin2022pwr,seurin2024assessment}. At each step, the agents were selecting fresh-, once-, and twice-burned Fuel Assemblies (FA)s to place in a reactor core. The reward was given by an objective function the reactor designer aimed to maximize. It encapsulated minimizing the LCOE and the sum of penalty terms incurred when safety-related constraints were not satisfied:
 \begin{equation}
     \max_{x} f(x) = - LCOE (x) - \sum_{i \in C} \gamma_i \Phi(x_i) + 1 \times \delta_{\forall i, x_i \le c_i}
    \label{eq:NucCoreObj}
\end{equation}
where $C = \{ (c_{i})_{i \in \{1,..,7\}} \}$ is the set of constraints and $(\gamma_{i})_{i \in \{1,..,7\}}$ is a set of weights attributed to each constraint, which optimal values were found to be equal to 25,000 from a sensitivity study conducted in \cite{seurin2024assessment} for the range [1,000-100,000]). If $ (x_{i})_{i \in \{1,..,7\}}$ are the values reached for the core ($x$) for the corresponding constraints $(c_{i})_{i \in \{1,..,7\}}$, $\Phi(x_i) = \delta_{x_i \le c_i} (\frac{x_i - c_i}{c_i})^2$, where $\delta$ is the Kronecker delta function. The formulation of $\Phi(x_i)$ serves as a sort of distance to the feasible space. Therefore, if a new LP constructed did not satisfy the constraints, the agents were punished with greater penalties. As a result, the goal of the agents were to choose the optimal fuel arrangement in the core which maximized the objective function. Within a day, the agents managed to learn to always find feasible patterns and find ones with very low LCOE. Table \ref{tab:cosntraintsforthepwr} displays the constraints and target values considered during these studies:

 \begin{table}[H]
     \centering
     \caption{Constraints for the PWR LP optimization problem \cite{seurin2022pwr,seurin2024assessment}. There are two and three choices for IFBA$^{\star}$ and WABA$^{\star}$, respectively: 128 and 156. and 0, 12, or 24. }
    \begin{tabular}{c|c|c|c|c|c|c|c}
     \hline
   FOMs & $L_{cy}$ [EFPD] & $F_{\Delta h}$ [-] & $F_q$ [-] & $C_b$ [ppm] & 
   \makecell{Peak Bu \\ $^{\star \star}$[GWd/tHm]} &  \makecell{\# \\ enrichment} & \makecell{\# \\ IFBA} \\
     \hline
    Constraints     & 500 & 1.45 & 1.85 & 1200 & 62 & [2,3] & [1,3]\\
      Inequality    & $\ge$ & $\le$ & $\le$& $\le$& $\le$ & $ \ge \le$ & $\ge \le$ \\
     \end{tabular}
     \label{tab:cosntraintsforthepwr}
     \begin{tablenotes}
        \item $^{\star}$ Integral Fuel Burnable Absorber (IFBA) is a type of Burnable Poison (BP). The other one considered in this study is the Wet Annular Burnable Absorber (WABA).
        \item $^{\star \star}$ This limit refers to the maximum burnup per pellet.
     \end{tablenotes}
 \end{table}
 
Moreover, we empirically demonstrated on the PWR but also classical benchmarks from the cec17 test suite \cite{awad2017problem}, that RL is optimally leveraged when it is applied similar to a Gaussian Process where the acquisition function is learned via a policy, which then converges close to the optimum solution after a certain number of steps \cite{seurin2024assessment}. This method, which we coined \textit{automatic supervised setting} in Section \ref{sec:introduction}, is similar to \textit{the contextual bandit} problem in RL.

\subsection{SO algorithms}
\label{sec:stochasticoptimization}
The performance of the RL algorithms will be measured against classical SO algorithms. The heuristics and computational approaches from SO's methodologies could be classified between local search (e.g. SA \cite{kirkpatrick1983optimization}, TS \cite{glover1989tabu}...), evolutionary (e.g., GA \cite{holland1992genetic}, ES \cite{beyer2002evolution}...), Swarm (e.g. Ant Colony Optimization, Particle Swarm Optimization \cite{lin2014themaxmin,li2022areview}), and hybrid or \textit{ensemble} (e.g., Teach Learning based Genetic Algorithm (HTLGA) \cite{li2022areview}, Prioritized replay Evolutionary and Swarm Algorithm (PESA) \cite{radaideh2022pesa}), which combined the large coverage potential of an evolutionary or swarm (i.e., global search) with a local search method used as a meta-heuristic to choose hyper-parameter or merely as a local improvement. 

We will leverage and describe SA, GA, and a novel parallel TS in Sections \ref{sec:simulatedannealing}, \ref{sec:geneticalgorithm}, and \ref{sec:tabusearch}, respectively. The latter are chosen because they are the most widely used methodology for solving the core loading optimization problem \cite{kropaczek2019largescale}. We will also utilize and describe in Section \ref{sec:prioritizedexperiencereplay}, the state-of-the-art ensemble method PESA, which was applied in the field of Boiling Water Reactor fuel bundle optimization \cite{radaideh2022pesa}. 

\subsubsection{Simulated Annealing (SA)}
\label{sec:simulatedannealing}

 Simulated annealing \cite{kirkpatrick1983optimization} is a general-purpose local search heuristic widely used in the optimization of the Light Water Reactor (LWR) core LP. This method is inspired by the process of cooling materials in the search of a metallic lattice structure with optimal configuration. This optimum corresponds to a state of minimum energy in the physical or the minimum objective function in the optimization process. At each step of the sequential SA, a solution is generated from the neighbor $\mathcal{N}(x)$ of the current candidate $x$. $\mathcal{N}(x)$ is delineated by user-defined perturbation operators (e.g., all solutions that can be reached by perturbing one entry of $x$). The objective function of the new solution is compared to the current one by evaluating $dE = E_{current} - E_{new} = E(x) - E(\mathcal{N}(x))$. If $dE$ is positive then the new solution is better hence it is accepted otherwise it can be accepted with probability $\rho \sim e^{\frac{dE}{T}} = e^{\frac{E(x) - E(\mathcal{N}(x))}{T}}$, where $T$ is called the annealing temperature, which classically evolves between $T_{max}$ and $T_{min}$ and help to avoid being stuck in a sub-optimal area of the objective space by allowing uphill moves. There, a higher temperature allows more states to exist and therefore increases the \textit{diversification} (equivalent to the \textit{exploration} in RL). Close to $T_{min}$, only downhill moves are accepted, increasing the \textit{intensification} (equivalent to the \textit{exploitation} in RL). This process is similar to the process of Markov Chain Monte Carlo (MCMC) in statistics \cite{bertsimas2008introduction}.
 
 Based on our literature review, it appeared that the PSA (PSA) with mixing of states \cite{chu1999parallel} is the state-of-the-art method applied to the core LP optimization \cite{kropaczek2009copernicus,ottinger2015bwropt}
alongside the lam cooling schedule  \cite{kropaczek2009copernicus}. There, the temperature does not evolve between $T_{max}$ and $T_{min}$ but is statistically derived from all parallel chains, and the optimization stops when the acceptance rate falls below a threshold \code{$\rho$}, equal to 10\% and 0\% in \cite{kropaczek2009copernicus} and \cite{ottinger2015bwropt}, respectively. Additionally, each processor is marching down its own Markov Chain but they are sharing the same annealing temperature, which is calculated based on a statistic derived from all chains as follows: At the end of the $i^{th}$ temperature evaluation controlled by the parameter \code{chain\_size}
, the starting solution of each chain is sampled with a certain probability 
from a pool of available candidates (see the corresponding literature for the sampling and cooling rate \cite{kropaczek2009copernicus}):
\begin{equation}
    \rho_i^l = \frac{e^{-\frac{E_i^l}{T_i}}}{\sum_{k = 1}^{\code{nChain}} e^{-\frac{E_i^k}{T_i}}}
    \label{eq:sasampling}
\end{equation}
where $\rho_i^l$ is the probability of starting a chain with the best solution ever found by chain $l$ after completing $i$ steps, \code{nChain} (equal to \code{ncores} in our case) is the number of Markov Chain operating in parallel, $E_i^k$ is the objective function of the last accepted state of chain $k$ at step $i$, and $T_i$ is the annealing temperature which follows the annealing schedule described by equation \ref{eq:lamannealing}: 
\begin{equation}
    \frac{1}{T_{i+1}} = \frac{1}{T_{i}} + \lambda \frac{T_i^2}{\sigma_i^2} \frac{1}{\sigma_i} f(\rho_i) 
    \label{eq:lamannealing}
\end{equation}
where $\sigma_i$ is the standard deviation of the objective function of all the accepted solution at step i, $\rho_i$ is the acceptance rate at step i, $T_0 = \alpha \sigma_0$, which is evaluated with a warm-up batch of candidate solutions, \code{$\alpha$} and \code{$\lambda$} are the initialization and quality factors, respectively, usually between 1 and 2, and  $f(\rho) = \frac{4\rho(1 - \rho)^2}{(2 - \rho)^2}$. Here, \code{$\alpha$} and \code{$\lambda$} are hyper-parameters, where \code{$\lambda$} controls the cooling rate and hence the speed at which the algorithm converges.
On top of that, at each step, an entry of the action is perturbed with probability \code{$\chi$}. Lastly, we need to define a move or a neighbour $\mathcal{N}(x)$  in each Markov Chain. To adapt to the problem defined in \cite{seurin2022pwr,seurin2024assessment}, at each step, an entry of the input vector is perturbed between a lower and upper bound with probability \code{$\chi$} (see Table \ref{tab:psahyperparameters} for a study of its influence on the quality of the LPs found).
\subsubsection{Genetic Algorithm (GA)}
\label{sec:geneticalgorithm}
GA-based algorithms \cite{holland1992genetic,beyer2002evolution,goldberg1989genetic} are inspired by the Darwinian theory of natural selection. An initial population of candidate solutions is generated and evolves generation after generation by experiencing a \textit{selection} of individuals with a bias toward the survival of the fittest, who will mate via \textit{crossovers} to generate off-springs for the next generation, themselves undergoing \textit{mutation}. The goal of the mutation is to help increase the diversity of the solution found, avoid falling in bad local optima, and delay the so-called \textit{genetic drift}. GA is naturally parallelizable as each individual in the population can be evaluated individually. Most efforts in the LP optimization literature and in the nuclear community in general, therefore, has been to apply various selection, crossover, and mutation techniques to each specific problem \cite{li2022areview}.

The selection operator is usually based on elitism for which the individual's selection probability is proportional to a fitness function with roulette \cite{castillo2014comparison,israeli2014novel} or linear ranking \cite{israeli2014novel}. Tournament selection has also been leveraged to increase diversity \cite{garco2008new}, for which two individuals are randomly selected and the one with the highest fitness is kept. Diversity can also be enhanced by encapsulating individuals in different niches, which fitness depends on the fitness of every candidate in the same niche \cite{pereira2008aparallel}.  Then, additional efforts were steered toward developing appropriate crossover and mutation operators that conserves fuel inventory \cite{parks1996multiobjective,erdogan2003apwr,delcampo2004development,garco2008new,israeli2014novel} or add more diversification to the crossover operator \cite{israeli2014novel}, with sometimes varying or adaptive mutation rate \cite{israeli2014novel,shaukat2010optimization,garco2008new}. The multiprocessing naturally ensues from the GA paradigm, in which each individual can be evaluated in parallel (which is the most computing-intensive process). However, \cite{pereira2008aparallel} studied more advanced parallelism with a combination of island GA and Niching techniques.

It was therefore hard to crystallize on a single GA version. For this work, the Evolutionary (\code{$\mu$},\code{$\lambda$}) Strategy (ES) \cite{beyer2002evolution}
is adopted. ES is a successful version of GA applied in the field of Boiling Water Reactor (BWR) bundle design \cite{radaideh2021large}. In ES, on top of the classical input in GA, a strategy vector $s$ must be initialized, which is utilized at the mutation stage (see below). 

There, initially, \code{$\lambda$} individuals are generated. At the end of a generation (i.e., \code{$\lambda$} evaluations), the best \code{$\mu$} individuals out of the \code{$\lambda$} are selected based on their fitness $F(y_i)_{i \in \{1,...,\mu\}}$. \code{$\lambda$} represents the number of individuals and hence the number of operations on the \code{$\mu$} individuals (without replacement) to generate enough offspring for the next generation. \code{$\mu$} could be low to impose more \textit{intensification} \cite{parks1996multiobjective,erdogan2003apwr} when generating the \code{$\lambda$} individuals of the next population. After selection, the \code{$\mu$} individuals will either experience crossover (with probability \code{cxpb}), mutation (with probability \code{mutpb}) but not both (hence \code{cxpb} + \code{mutpb} $\le 1$), or will be cloned and appears in the next generation. 
The global crossover and mutation probability are constant but the mutation strength of each attribute (probability of mutating an entry of an individual $y_i$) evolves during the search via a strategy vector $s_i$. Hence each individual is characterized by $\tilde{y_i} = (y_i,s_i,F(y_i))$ \cite{radaideh2022pesa}. 

If crossover, two random individuals ind1 and ind2 are chosen, and two locations pt1 and pt2 are sampled. These two individuals then exchange the portion of their input defined by [pt1,pt2] and the resulting offspring is the new ind1. If mutation, an individual is randomly selected and mutated according to the log-normal $\sigma$-adaptation scheme as applied in \cite{radaideh2022pesa}. For each entry of the input vector, the probability of mutation is tuned as follows (we show here the update for a continuous space and refer the reader to the appropriate references for the integer space for sake of brevity): 
let $\tau = \frac{1}{\sqrt{2 |y_i|}}$ and $\tau^{\star} = \frac{1}{\sqrt{2 \sqrt{|y_i|}}}$: \\
\begin{equation}
\tilde{y_i} \leftarrow \{ \begin{array}{l}
    s_{ik} \leftarrow s_{ik} e^{\tau g + \tau^{\star} g^{\star}}\\
    y^{new}_{ik} \leftarrow y_{ik} + s_{ik} \mathcal{N}(0,1) \\
    f(y_i) \leftarrow f(y^{new}_i)
  \end{array} 
\end{equation} 
where $g$ and $g^{\star}$ are randomly sampled $\sim \mathcal{N}(0,1)$, $y_{ik}$ and $s_{ik}$ are the kth entry of $y_i$ and $s_i$, respectively. $s_i$, the strategy vector or the mutation strength, is self-controlled (while in classical GA the probability of mutation of an entry \code{indpb} is usually fixed) because the new individual generated will be selected depending on the fitness obtained after applying the new strategy. The strategy vector must also be re-calibrated such that it remains between user-defined upper and lower bounds. Lastly, if cloning happens, a randomly chosen element is copied to the new batch of off-springs. 

\subsubsection{Tabu Search (TS)}
\label{sec:tabusearch}
Tabu Search (TS) \cite{glover1989tabu} is another popular algorithm leveraged within the field of fuel LP optimization \cite{li2022areview,castillo2004bwr,castillo2011huitzoctli}. By introducing a concept of memory, it tries to imitate intelligent human behavior during the search process, which is often used as a meta-heuristic to guide lower-level solvers \cite{fiechter1994aparallel}. 

Similar to SA, new solutions are obtained from the neighborhood of $x$, $\mathcal{N}(x)$, by perturbing some entries of the candidate solutions vector $x$. The operation can result in a swap of two entries of the input vector in the case of a CO problem or a perturbation between an upper and lower bound in the case of integer and continuous optimization. In TS, however, the whole or part of the neighborhood of the candidate vector is evaluated first before deciding which solution to move on with. By contrast in SA, the candidate vector experiences most of the time one perturbation only and the acceptance of the resulting solution is based on a metropolis criteria (see section \ref{sec:simulatedannealing}). It then proceeds to the best non-tabu (see below at \textit{short-term memory} for its meaning) solution obtained. Nevertheless, when the number of possible perturbations is too large, only a sample of them could be performed until a better solution is found as experimented in large-scale constrained BWR LP optimization problems (10\% of the neighborhood in \cite{castillo2004bwr}). 

The major characteristics of a TS algorithm are listed as follows:
\begin{enumerate}
    \item The first is the \textit{type of perturbation}. It can be either swapping two entries of the input vector (or even two sets of swap \cite{gu2021reinforcement}), which is adapted to CO for which the input embedding is composed of unique labeled entries, or perturbation of an entry between a lower and upper bound. For the former, $N^2$ actions are available, while N for the latter, where N is the number of entries in an input vector. 
    \item The second element is the so-called \textit{short-term memory}, which prevents moves that has been performed to be revisited during a user-defined period called \code{tabu\_tenure} (usually between 6 to 14 \cite{glover1989tabu}), preventing cycling to already visited solutions. However, an \textit{aspiration criterion} may accept tabu solutions if they have shown to improve the best solution ever found.
    \item To increase diversification, a \textit{long-term memory} is implemented via a frequency penalization \code{penalization\_weight}, which fosters moves in un-visited areas of the search space. 
    \item To increase intensification, an intermediate or \textit{medium-term memory} is sometimes leverages to increase the frequency of moves in regions that contain good solutions \cite{fiechter1994aparallel,hill2015pressurized} (e.g., good moves that were not selected during the visitation of a neighborhood). 
    \item Lastly, \textit{stopping criteria} must be defined when the search stops, which can be defined as the number of moves without improvement \cite{castillo2004bwr}. 
\end{enumerate}

The literature on utilizing this technique is less rich than SA and GA but resulted in significant contributions on BWR \cite{castillo2004bwr,castillo2011huitzoctli} and PWR \cite{lin2014themaxmin,wu2016quantum,hill2015pressurized} optimization. 

However, to the knowledge of the authors of this work, no parallel versions of TS were implemented to solve the PWR or BWR LP optimization problems but must be here for TS to be a valid method to compare RL and other SO with. There are different possibilities to implement parallelism in TS \cite{bortfeldt2003aparallel}: 
\begin{enumerate}
\item 
 Parallelization of operation within a stage, for instance, evaluating all available moves in parallel. 
\item 
 Allocate different processors to solve different independent subproblems and recombine them into the input vector.
\item 
 Create multiple processing chains that operate in parallel and interact.
\end{enumerate}Here, the third option was considered due to ease and its resemblance to the PSA implementation: \code{ncores} Markov Chains evolve in parallel and interact at the end of \code{chain\_size} operations. To restart the chain, we propose four different sampling strategies: The first is a \code{hard} restart, where each chain restarts with the best solution ever found as showcased in equation \ref{eq:hard}. The next two are \code{roulette} and \code{rank} (i.e., using a linear ranking (LR)) given in equations \ref{eq:roulette} and \ref{eq:linearranking}, respectively, which are classically used in GA for selection strategy \cite{israeli2014novel}.  The last one, called \code{softmax}, is similar to the parallel SA, where a hyper-parameter \code{$\kappa$}, the inverse temperature in ML literature, has been added for the function to resemble a softmax policy as given in equation \ref{eq:softmax}. The choice is controlled by the hyper-parameter \code{reinforce\_best} in the Neuro-EvOlution with Reinforcement Learning (NEORL) package \cite{radaideh2023NEORL}:
\begin{equation}
 p_i = \{ \begin{array}{l}
   1 \quad \text{if} \quad E_i \in argmax\{E_k, k\in\{1,...,\code{ncores}\}\} \\
   0 \quad \text{otherwise} \\
  \end{array}  
  \label{eq:hard}
\end{equation}
where $p_i$ is the probability for selecting sample $i$ to restart a chain
\begin{equation}
p_i = \frac{E_i}{\sum_k E_k}
\label{eq:roulette}
\end{equation}
\begin{equation}
p_i = \frac{1}{n}(2 - m + 2 (m - 1)\frac{rank_i - 1}{n - 1})
\label{eq:linearranking}
\end{equation}
where $rank_i$ is the rank of the solution found compared to all chains and \code{m} is a hyper-parameter that controls the magnitude of the differences between the $p_i$.
\begin{equation}
p_i = \frac{exp(-\kappa E_i)}{\sum_kexp(-\kappa E_k)}
\label{eq:softmax}
\end{equation}

When the $i^{th}$ element is chosen to restart the $p^{th}$ chain, the best energy and solution are transferred like in SA. The Tabu heuristics \code{tabu\_structure} (known frequency of certain moves, \textit{short-} and \textit{long-term memory}, etc...) is however shared across the chains. Lastly, we utilize a parameter \code{$\chi$} which is the probability of perturbing an entry of the input vector. Similar to \cite{castillo2004bwr}, if none of the moves sampled improve the best individual found by the chain, the remainder of the entries are sampled as well. We however do not stop at the first entry which improves the individual but the best of all that does.

This implementation as well as other sampling mechanisms will be available in NEORL \cite{radaideh2023NEORL} after publication for reproducibility. 

\subsubsection{Prioritized experience replay for parallel hybrid
evolutionary and swarm algorithms (PESA)}
\label{sec:prioritizedexperiencereplay}
Prioritized experience replay for parallel hybrid evolutionary and swarm algorithm (PESA) \cite{radaideh2022pesa} is a new type of neuroevolution algorithm that leveraged the benefit of different heuristics algorithms namely ES, SA, and Particle Swarm Optimization (PSO) \cite{kennedy1995particle}. ES was described in \ref{sec:geneticalgorithm} and SA in \ref{sec:simulatedannealing} but PSO is introduced for the first time. This algorithm is inspired from the flock of birds or fishes and belongs to the class of swarm algorithm. Each member of the swarm at time $t$ is characterized by a position $x_i^t$ and a velocity $v_i^t$. The position is updated as $x_i^{t+1} = x_i^t + v_i^{t+1}$ where the update of the velocity depends on the knowledge of the best solution found by the $i^{th}$ member $pbest_i^t$, but also the best solution overall $best^t$: $v_i^{t+1} = f_{pso}(v_i^t,pbest_i^t - x_i^t,best^t - x_i^t)$. The function $f_{pso}$ depends on the different version of PSO, and the reader should refer to \cite{radaideh2022pesa} for its description.
To reduce the burden of hyperparameters tuning, the parallelism is such that the number of chain $\times$ the length of the chain in SA (i.e., $\code{ncores} \times \code{chain\_size}$) is equal to the number of individuals in ES (i.e., \code{$\lambda$\_}) and PSO populations (i.e., \code{npar}). As a result, each algorithm finish their inner loop at the same time, reducing delays and coupling between the methods.

While the the three optimization algorithms are progressing, they store their solutions in a shared buffer, which they can draw sample from frequently in their inner loops (e.g., to restart a Markov chain in parallel SA or injecting new individual in the populations of ES and PSO) based on the notion of prioritized experience replay in deep RL \cite{mnih2015human} described in equation \ref{eq:prioritizedexperiencereplay} . As a result each algorithm shares data on-the-fly, which resulted in superiority on many classical benchmarks against their standalone counterparts in \cite{radaideh2022pesa}.

\begin{equation}
    p_i = \frac{1}{rank(i)}, \quad \rho_i = \frac{p_i^{\alpha}}{\sum_k^{\kappa}p_k^{\alpha}}
\label{eq:prioritizedexperiencereplay}
\end{equation}
where $rank(i)$ is the rank of the $i^{th}$ solution in the buffer. $\rho_i$ is the probability of selecting the $i^{th}$ sample, $\kappa$ is the size of the buffer and $\alpha \in [0,1]$ is called the priority coefficient. 

\subsection{Input decision variables for SO}
\label{sec:inputdecision}

For the sake of implementation simplicity, we opted to use a similar input format for both RL and SO methods. Given that our SO algorithms also handle integer decision variables, we can utilize elements of the action space from RL as input vectors for SO. The decoding process for candidate inputs ensures the conservation of fuel inventory and basic heuristics, such as leaving twice-burned FAs at the periphery. This encoding approach eliminates the need for specialized crossover or mutation preprocessing steps. Unlike many existing papers, our input does not represent an ensemble of FAs to shuffle, but rather a sequence of assemblies to choose (for fresh fuels) and/or shuffle (for fresh-, once-, and twice-burned FAs). This sequential approach ensures the manageability of the fuel inventory: core locations are traversed sequentially, and if an unavailable FA is selected for a particular location, a random alternative assembly is chosen.

Traditionally, in TS, a move represents a binary swap when applied to core design. However, in our case, we randomly perturb one entry within the upper and lower bounds defined by the action space. This is facilitated by employing the \code{perturb} for \code{swap\_mode} in NEORL for continuous and mixed-integer optimization. The objective function used is similar to the reward defined in Equation \ref{eq:NucCoreObj}, aiming to maximize the LCOE while adhering to seven key safety and operational constraints.

The evaluation of the population poses a computational bottleneck due to the core physics code's (SIMULATE3) computational cost. Thus, parallelism is implemented during the calculation of the objective function for all candidates discovered during the search. In RL, each agent performs its rollout individually, while for TS and SA, each Markov Chain is assigned its processor, and in ES, each individual in the population is allocated one.  For PESA, a third of the processor is assigned to each heuristic (i.e., SA, ES, and PSO). We utilize 32 cores for each algorithm except PESA for which we use 30 cores (e.g., the populations of ES and PSO are comprised of 10 elements and there are 10 Markov chains in SA). 

To avoid biasing the input structure in favor of the RL paradigm, we briefly explore optimal hyperparameters for the SO-based approaches (except for TS) in Appendix \ref{appendix:hyperparameterforpsaandes} and assign this optimal ones to the SA and ES subroutine of PESA in NEORL.

\section{Test sets, objectives, and constraints}
\label{sec:testsets}

\subsection{Classical PWR benchmark}
\label{sec:classicalpwrbenchmark}

To demonstrate the performance of the RL algorithms in our prior works \cite{seurin2024assessment,seurin2024multiobjective}, the original problem studied is the optimization of the equilibrium cycle of a 4-loop Westinghouse-type reactor with 193 assemblies, 89 of which are fresh (including the center assembly). This means that 88
will remain in the reactor (but potentially at a different positions) for the
next operational cycle after refueling, while 16 will remain for three cycles
in total. Note that due to symmetry, the center assembly could stay in the reactor core for the next cycle. However, because we are operating in equilibrium cycle (i.e., we use the same pattern across all operational periods), we must refuel the center assembly at the beginning of cycle.  

To reduce the search space, mirror one-eighth symmetry boundary conditions were applied. The objective to optimize is given by equation \ref{eq:NucCoreObj} and the composition of the potential different FAs to chose from is given in Table \ref{tab:examplesofthecontent}. Moreover, to help the agent learn, we have chosen 24 fuel assemblies out of the possible 36 combinations of enrichment and BPs to avoid ones that did not make sense (e.g., 24-WABA and 156-IFBA with an Uranium enrichment of 4.00\%). Lastly, as explicated in \cite{seurin2022pwr,seurin2024assessment,seurin2024multiobjective}, we applied expert tactics to reduce the search space, which is comprised of about $10^{29}$ candidate solutions. The few tactics include putting twice-burned fuel at the periphery and preventing fresh ones from being located there, enforcing a ring of fresh FAs between the in-board and the periphery, and preventing squares of fresh FAs in the core. 
\begin{table}[H]
    \centering
    \caption{Examples of the content of FAs the agents can possibly utilize to refuel the reactor. There is a discrete choice from the values provided.}
    \begin{tabular}{c|c|c}
    \hline
    Enrichment [\%] & IFBA [-] & WABA [-]\\
    \hline
      $\{4.00,4.20,4.40,4.60,4.80,4.95\}$   &  $\{128,156\}$ & $\{0,12,24\}$\\
    \end{tabular}
    \label{tab:examplesofthecontent}
\end{table}

\subsection{Extended PWR designs}
\label{sec:extendeddesigns}
The LP optimization problem is studied with the goal to find the best fuel arrangement that yield economic and yet safe operation perspectives. This problem, characterized by numerous trade-offs, has been a focus of research for over six decades, initially addressed by human experts and extensively optimized for specific reactor designs using expert knowledge \cite{seurin2022pwr,seurin2024assessment}. As a result, when optimizing, researchers are making lots of conservative assumptions, such as fixing the number of fresh FAs loaded. 

Our extended designs studied therefore will compare the methods on more intricate scenarios with less fresh fuels. The first new scenario involves an extension of the 89-eighth symmetry but with quarter symmetry and the same type of fresh fuels, which complexity amounts to $10^{34}$ possible solutions. The second involves reducing the number of fresh fuels from 89 to 85 (thus removing one FA per quadrant). It is studied with quarter symmetry since achieving an eighth-symmetry problem is not feasible in this case. The complexity of this problem is approximately $10^{38}$. 
The other two problems involve core optimization with 81 fresh FAs (i.e., two assemblies removed per quadrant) and are considered with eighth- and quarter symmetries. The complexity of these problems amounts to $10^{28}$ for eighth symmetry and $10^{31}$ for quarter symmetry. 

 To compensate for the loss of reactivity in the 85 and 81 cases, which can lead to a reduction in $L_{cy}$, we anticipate that the average feed enrichment will need to be increased. However, this adjustment may also result in higher peaking factors ($F_{\Delta h}$) and peak $C_b$. The latter can be compensated by a high WABA but the former is more tricky. Therefore, these problems are significantly harder to solve (even for the 81 cases in spite of being smaller in size). In practice, for a 4-loop PWR with core power density of about 110 kW/L, operating an 18 month cycle, core designer have been persisting with 89 or sometimes 85 fresh FAs in their feed. Demonstrating feasibility in the 81 cases could therefore represent a significant stride toward attaining economic benefits beyond expert knowledge. Feasibility was achieved in our concurrent paper \cite{seurin2024physics}, but the goal of this work is to compare the performance of the various methods and not necessarily find a feasible solution.

\subsection{Objectives and constraints}
\label{sec:objectivesandconstraints}

In \cite{seurin2024assessment} ,  we did not impose an upper limit on $L_{cy}$ as showcased in Table \ref{tab:cosntraintsforthepwr}. Nevertheless, an 18-month campaign for a typical 4-loop Westinghouse-type reactor amount to about $1.5\times 365.25 = 547.9$ days. With a availability factor of 98\%, it reduced to 537 days. With about 30-40 days-long outages, we can expect the NPP to run around 500 days before shut-down. Compared to our other study in \cite{seurin2024assessment}, where we set this 500 as lower limit, here we target 500 EFPD exactly.

For all these cases, the same type of fuel loaded for the earlier studies \cite{seurin2024assessment,seurin2024multiobjective} will be utilized (see Table \ref{tab:examplesofthecontent}). All the constraints of Table \ref{tab:cosntraintsforthepwr} are considered and we add the following as discussed above: $500 \ge L_{cy}$. As a result, the constraint is now $L_{cy} = 500$. 


\section{Results and Analysis}
\label{sec:resultsandanalysis}

\subsection{PPO versus legacy algorithms on  (single-objective optimization)}
\label{sec:ppoversuslegacyextended}
As we argued in \cite{seurin2024assessment}, that we implemented RL as a Gaussian Process (GP), where the acquisition function is replaced by a policy that converges near the optimal solution. In this setup, agents at each step seek the steepest ascent direction based on their understanding of the objective space’s curvature, while SO-based methods take random steps to navigate the search space. Therefore, it’s accurate to say that while agents in SO follow a guided random walk, employing a heuristic in their decision-making process, in PPO, a policy generates solutions, and a critic learns and evaluates their quality. In PPO, the direction of best LP improvement, i.e., the heuristic, is learned.

In order to explore if we can enhance the robustness, stability, and quality of solutions by learning these heuristics, we conducted experiments with SO approaches on the problems outlined in Section \ref{sec:testsets}. The study of the best hyperparameters for SO is detailed in Appendix \ref{appendix:hyperparameterforpsaandes} and we used the same parameters as in \cite{seurin2024assessment} for PPO. For PESA, we used the best hyperparameters of ES and SA for their respective modules and the default values for the PSO module. Each algorithm was run with up to a maximum of 20,000 samples for 10 experiments, utilizing 32 cores (i.e., 32 agents) on an Intel Xeon Gold 5220R clocked at 2.20 GHz.

\subsubsection{Eighth symmetries}
\label{sec:eighthsymmetriesstats}
 Figures \ref{fig:89eighthcases} and \ref{fig:81eighthcases} in appendix \ref{appendix:evolutionoftheaverage} illustrate the evolution of the mean score (blue curve with y-axis on the left) and the maximum objective (i.e., best LP, red curve with y-axis on the right) per generation (normalized to show 200 generations in total for proper visualization) in the 89- and 81-eighth cases, respectively. The figure \ref{fig:89eighthcases}.(6) and \ref{fig:81eighthcases}.(6) showcase the evolution of the best objective for the last 160 generations for all cases in the same figures for ease of comparison. 
 
 Despite the smooth improvement in all rewards, PPO consistently outperforms the other four methods. In particular, PPO identifies the best pattern and achieves the highest mean reward per episode by the last generation (refer to Tables \ref{tab:89eighthcasestats} and \ref{tab:81eighthcasestats}). It's worth noting that the slight flattening of the curve for PPO in later generations is due to the scale on the y-axis, and if we could zoom in, we would observe that PPO continues to learn effectively. Approximately after 25 generations, PPO's average reward remains higher than that of TS and PSA, and after about 100 generations, both the average and best patterns found by PPO surpass those of ES. Furthermore, PPO exhibits the highest mean and lowest standard deviation across the 10 experiments throughout all generations and for both problems. This indicates greater robustness and stability in PPO's performance compared to the other methods. Additionally, we observe the learned rule of thumb over time, as the reward steadily increases with a small standard deviation. This suggests that PPO consistently generates better or close-to-better solutions across generations, which could be the case for TS and PSA as well but to a smaller extent. For PPO, the reward is recorded for every pattern generated. In contrast, for TS, PSA, and PESA the reward is recorded at each function evaluation, even if the solution is not accepted. For ES, the reward is recorded for each member of the population, even if they are not selected for the next generation, which explains why the average reward per generation (blue curves) of PPO is so much better than the other algorithms.

ES proves to be the second-best algorithm for the type of problem we are studying, although it fails to find a feasible pattern within the allotted 20,000 samples on average. PSA closely resembles TS. Interestingly, both are local search algorithms \cite{li2022areview} and slowly perturb solutions at each step. While TS accepts the best non-tabu move (i.e., the perturbation of one entry), PSA perturbs an average of three entries (with a parameter \code{$\chi$}=0.1) at each step until no solution is accepted as explained in Section \ref{sec:stochasticoptimization}. Although they find similar patterns on average, PSA exhibits less certainty with a higher standard deviation and a lower mean reward. Our problem thus appears to be more amenable to global search methods for quickly finding reasonable solutions (e.g., PPO, ES, and PESA), while subsequent refinement of solutions requires more exploitation with smaller changes (e.g., PPO). RL achieves this automatically through policy refinement and agent adaptation to the curvature of the loss function, outperforming the random crossovers and mutations of ES in this scenario. We would have hoped that PESA would behave as PPO since it is leveraging the best of both worlds (i.e. global and local search), noting its superior performance in \cite{radaideh2022pesa}. However, PESA's performance relies upon the interaction between PSO, SA, and ES. The intricacies between the hyper-parameters of SA and ES when they are combined, as well as their relationship with the PSO module must be further investigated to improve its performance in the PWR cases. Lastly, as explained in Section \ref{sec:inputdecision}, only a third of the processor is assigned to each sub-heuristic module hence the performance of ES and SA stand-alone above are obtained with populations three times larger, affecting their performance tremendously. Access to larger number of processors would be beneficial for PESA but also PPO and ES. We left a study with a greater number of processors for future research endeavors.


\begin{table}[H]
    \centering
    \caption{Best and average objectives, and standard deviation $\sigma$ over the experiments for the best pattern (left) and average pattern in one generation (right)  at the last generation. 89-eighth case}
     \begin{minipage}[b]{0.45\linewidth}
    \begin{tabular}{c|c|c}
   Cases Obj & Avg Max & Final $\sigma$ \\
 \hline

TS& -67.273 & 39.533 \\
ES& -27.9809  & 23.894 \\
PSA&  -70.232 & 53.953 \\
PESA &-36.600 &  31.118 \\
PPO & \textcolor{red}{-8.124} &  \textcolor{red}{5.872}\\
    \end{tabular}
    \end{minipage}
     \begin{minipage}[b]{0.45\linewidth}
     \begin{tabular}{c|c|c|c}
    Cases & Max reward & Avg reward & Final $\sigma$ \\
    \hline
TS& -550.3667 & -1119.032 & 467.767 \\
ES& -290.208& -676.408 & 327.979 \\
PSA& -1183.895 & -2009.962 & 986.165 \\
PESA &  -2244.878 & -4169.377 & 992.544 \\
PPO & \textcolor{red}{-22.102} & \textcolor{red}{-58.850} & \textcolor{red}{37.199} \\
     \end{tabular}
     \end{minipage}
    \label{tab:89eighthcasestats}
\end{table}

\begin{table}[H]
    
    \centering
    \caption{Best and average objectives, and standard deviation $\sigma$ over the experiments for the best pattern (left) and average pattern in one generation (right)  at the last generation. 81-eighth case}
     \begin{minipage}[b]{0.45\linewidth}
    \begin{tabular}{c|c|c}
    Cases & Avg Max & Final $\sigma$ \\
    \hline

TS & -90.611 & 44.600 \\
ES & -16.978 & 13.754 \\
PSA & -95.944 &  81.112 \\
PESA & -44.026 & 30.733 \\
PPO &  \textcolor{red}{-8.55} & \textcolor{red}{5.737} \\
    \end{tabular}
    \end{minipage}
     \begin{minipage}[b]{0.45\linewidth}
     \begin{tabular}{c|c|c|c}
    Cases & Max reward & Avg reward & Final $\sigma$ \\
    \hline
TS & -669.404 & -1116.359 &  388.178 \\
ES & -45.542 & -505.510 &  325.871 \\
PSA & -713.625 &  -1469.962 &  792.464 \\
PESA &-2869.270 & -3718.751 & 616.060 \\
PPO & \textcolor{red}{-14.299} & \textcolor{red}{-31.255} & \textcolor{red}{12.287} \\
     \end{tabular}
     \end{minipage}
    \label{tab:81eighthcasestats}
\end{table}

To ensure that PPO is statistically better than the other methods, we conducted statistical tests on the best patterns found. After running 10 experiments, the Friedman Tests (FT)s yielded successful results, indicating that at least one of the solutions is different from the others. Consequently, the Nemenyi post-hoc tests (NT)s can be computed, and the results are provided in Tables \ref{tab:nemenyitests89iehghtcase} and \ref{tab:nemenyitests81eighthcase}. All the NTs involving PPO (either in the last row or last column) succeeded, except against ES and TS in three and one case(s), respectively: mean reward at the last generation in the 89-eighth case, and for both in the 81-eighth case. mean reward at the last generation against TS in the 89-eight case. However, for the latter, the value is 0.11, which is close to $\alpha =0.1$, a valid threshold for significance \cite{seurin2024assessment,schlunz2016acomparative}. Moreover, ES is not statistically significantly better than TS, PSA, or even PESA in most cases. Hence, it's conceivable that by running 10 more experiments, PPO would likely become better than ES and TS. However, obtaining these additional samples in the SO cases takes more than a day, necessitating at least 80 more days in total of runs for marginal improvements. Therefore, we omit this study here for brevity. Nevertheless, we can still assert that PPO is superior in these types of LP problems.

\begin{table}[H]
    
    \centering
    \caption{Nemenyi Tests on the best LPs found: 89-eighth case. For the Friedman Test, the p is equal to 0.002 and 0.000 for the Max obj and reward, respectively. Highlighted in red are the successful tests with significance level $\alpha = 0.1$. Most tests for PPO succeeded.}
     \begin{minipage}[b]{0.45\linewidth}
    \begin{tabular}{c|c|c|c|c|c}
    Cases & TS & ES & PSA & PESA & PPO \\
    \hline\
TS  & 1.000& 0.522 & 0.900 & 0.522  & \textcolor{red}{0.001} \\
ES  & 0.522& 1.000 & 0.843 & 0.900  & \textcolor{red}{0.081} \\
PSA  & 0.900& 0.843 & 1.000 & 0.843  & \textcolor{red}{0.003} \\
PESA  & 0.522& 0.900 & 0.843 & 1.000  & \textcolor{red}{0.081} \\
PPO & \textcolor{red}{0.001}& \textcolor{red}{0.081} & \textcolor{red}{0.003} &\textcolor{red}{0.081}  & 1.000 \\
    \end{tabular}
    \end{minipage}
    \hspace{15pt}
     \begin{minipage}[b]{0.45\linewidth}
     \begin{tabular}{c|c|c|c|c}
      TS & ES & PSA & PESA &  PPO \\
    \hline
1.000& 0.900000  & 0.437&\textcolor{red}{0.010} &0.114 \\
0.900& 1.000000  & 0.210&\textcolor{red}{0.002} &0.275 \\
0.437& 0.210897  & 1.000&0.522 &\textcolor{red}{0.001} \\
\textcolor{red}{0.010}& \textcolor{red}{0.002198}  & 0.522&1.000 &\textcolor{red}{0.001} \\
0.114& 0.275862  & \textcolor{red}{0.001}&\textcolor{red}{0.001} &1.000 \\
     \end{tabular}
    
     \end{minipage}
    \label{tab:nemenyitests89iehghtcase}
\end{table}

\begin{table}[H]
    \centering
    \caption{Nemenyi Tests on the best LPs found: 81-eighth case.  For the Friedman Test, the p is equal to 0.000 for both the Max obj and reward. Highlighted in red are the successful tests with significance level $\alpha = 0.1$. Most tests for PPO succeeded.}
     \begin{minipage}[b]{0.45\linewidth}
    \begin{tabular}{c|c|c|c|c|c}
    Cases & TS & ES & PSA & PESA & PPO \\
    \hline
TS& 1.000 & \textcolor{red}{0.081} & 0.900& 0.683 &\textcolor{red}{0.001} \\
ES & \textcolor{red}{0.081} & 1.000 & \textcolor{red}{0.055}& 0.683 &0.522 \\
PSA & 0.900 & \textcolor{red}{0.055} & 1.000& 0.602 &\textcolor{red}{0.001} \\
PESA & 0.683 & 0.683 & 0.602& 1.000 &\textcolor{red}{0.037} \\
PPO & \textcolor{red}{0.001} & 0.522 & \textcolor{red}{0.001} & \textcolor{red}{0.037} &1.000 \\
    \end{tabular}
    \end{minipage}
    \hspace{15pt}
     \begin{minipage}[b]{0.45\linewidth}
     \begin{tabular}{c|c|c|c|c}
     TS & ES & PSA & PESA & PPO \\
    \hline
1.000& 0.351 & 0.900& 0.114& \textcolor{red}{0.010} \\
0.351& 1.000 & 0.114& \textcolor{red}{0.001}& 0.602 \\
0.900& 0.114 & 1.000& 0.351& \textcolor{red}{0.001} \\
0.114& \textcolor{red}{0.001} & 0.351& 1.000& \textcolor{red}{0.001} \\
\textcolor{red}{0.010}& 0.602 & \textcolor{red}{0.001}& \textcolor{red}{0.001}& 1.000 \\
     \end{tabular}
     \end{minipage}
    \label{tab:nemenyitests81eighthcase}
\end{table}

\subsubsection{Quarter symmetries}
\label{sec:quartersymmetriesstats}
In this section, we repeat the study we performed in Section \ref{sec:eighthsymmetriesstats} but on the more intricate cases in which we increases the symmetry from eighth to quarter.  Figures \ref{fig:89quartercases} through \ref{fig:81quartercases} in appendix \ref{appendix:evolutionoftheaverage} illustrate the evolution of the mean score (blue curve with y-axis on the left) and the maximum objective (i.e., best LP, red curve with y-axis on the right) per generation (normalized to show 200 generations in total for proper visualization) in the 89-, 85-, and 81-quarter scenarios. Similar to what was observed in Figures  \ref{fig:89eighthcases} and  \ref{fig:81eighthcases}, the average reward of PPO, TS, and PSA improve consistently but PPO does so much more rapidly. On the other hand, both ES and PESA experiences sharp improvement at the beginning and significant slow-down as soon as generation 50 for most scenarios, preventing them to surpass PPO in all but the 81-quarter scenario. The other discrepancies observed are not linear unlike in the comparison over the two eighth cases. Therefore we have separated the following discussion in three portions corresponding to each three quarter scenarios separately:

\paragraph{89-quarter scenario}: In light of the Tables \ref{tab:89quartercasestats}, PPO presents higher average objective and is therefore clearly superior. However, ES, and PESA found high-quality solutions. Consequently, the corresponding NTs failed (see Table \ref{tab:nemenyitests89quartercase}). Moreover, PPO has an average best reward high with a very low Final $\sigma$, meaning that it can find high-quality solutions with great certainty. This indicates again greater robustness and stability in PPO's performance compared to the other methods, which is a fundamental behavior for it to be accepted and deployed at Nuclear Power Plants (NPPs). 

\begin{table}[H]
    \centering
    \caption{Best and average objectives, and standard deviation $\sigma$ over the experiments for the best pattern (left) and average pattern in one generation (right)  at the last generation. 89-quarter case}
     \begin{minipage}[b]{0.45\linewidth}
    \begin{tabular}{c|c|c}
   Cases & Avg Max & Final $\sigma$ \\
 \hline

TS & -87.386& 51.677 \\
ES & -55.977&102.037\\
PSA & -139.152&246.855\\
PESA & -18.416& 12.777\\

\textcolor{red}{PPO} & \textcolor{red}{-5.575}& \textcolor{red}{0.584}\\
    \end{tabular}
    \end{minipage}
     \begin{minipage}[b]{0.45\linewidth}
     \begin{tabular}{c|c|c|c}
    Cases & Max reward & Avg reward & Final $\sigma$ \\
    \hline
TS&-241.316&-1127.546&806.814\\
ES&-123.691&-588.920&385.407\\
PSA&-942.399&-1793.312&1228.305\\
PESA&-2537.624&-4225.293&1813.520\\
\textcolor{red}{PPO}&\textcolor{red}{-35.328}&\textcolor{red}{-52.845}&\textcolor{red}{14.915}\\
     \end{tabular}
     \end{minipage}
    \label{tab:89quartercasestats}
\end{table}


\begin{table}[H]
    
    \centering
    \caption{Nemenyi Tests on the best LPs found: 89-quarter case. For the Friedman Test, the p is equal to 0.003 and 0.000 for the Max obj and reward, respectively. Highlighted in red are the successful tests with significance level $\alpha = 0.1$. Most tests for PPO succeeded.}
     \begin{minipage}[b]{0.45\linewidth}
    \begin{tabular}{c|c|c|c|c|c}
    Cases & TS & ES & PSA & PESA & PPO \\
    \hline\
TS&1.000&\textcolor{red}{0.010}&0.683&0.114&\textcolor{red}{0.001}\\
ES&\textcolor{red}{0.010}&1.000&0.275&0.900&0.763\\
PSA&0.683&0.275&1.000&0.763&\textcolor{red}{0.016}\\
PESA&0.114&0.900&0.763&1.000&0.275\\
PPO&\textcolor{red}{0.001}&0.763&\textcolor{red}{0.016}&0.275&1.000\\
    \end{tabular}
    \end{minipage}
    \hspace{15pt}
     \begin{minipage}[b]{0.45\linewidth}
     \begin{tabular}{c|c|c|c|c}
      TS & ES & PSA & PESA &  PPO \\
    \hline
1.000&0.683&0.900&0.114&\textcolor{red}{0.016}\\
0.683&1.000&0.351&\textcolor{red}{0.002}&0.351\\
0.900&0.351&1.000&0.351&0\textcolor{red}{.002}\\
0.114&\textcolor{red}{0.002}&0.351&1.000&\textcolor{red}{0.001}\\
\textcolor{red}{0.016}&0.351&\textcolor{red}{0.002}&\textcolor{red}{0.001}&1.000\\
     \end{tabular}
     \end{minipage}
    \label{tab:nemenyitests89quartercase}
\end{table}

\paragraph{85-quarter scenario}:
In this case, PPO still surpasses the other methods in terms of best pattern found, but the NTs failed against all but PSA (see Table \ref{tab:nemenyitests85quartercase}). TS exhibits a higher Avg Max and Max reward compared with PSA and managed to fail PPO's NTs against it. PESA exhibits an average behavior for the best pattern found on-par with ES this time, with an Avg Max of -20.978 and -20.744, respectively. While PPO is still superior, the differences are less important than for the three previous scenarios studied. 

\begin{table}[H]
    \centering
    \caption{Best and average objectives, and standard deviation $\sigma$ over the experiments for the best pattern (left) and average pattern in one generation (right)  at the last generation. 85-quarter case}
     \begin{minipage}[b]{0.45\linewidth}
    \begin{tabular}{c|c|c}
   Cases & Avg Max & Final $\sigma$ \\
 \hline
TS&-69.584&56.721\\
ES&-20.644&11.512\\
PSA&-113.199&69.750\\
PESA&-20.978&18.568\\

\textcolor{red}{PPO}&\textcolor{red}{-15.222}&\textcolor{red}{7.461}\\
    \end{tabular}
    \end{minipage}
     \begin{minipage}[b]{0.45\linewidth}
     \begin{tabular}{c|c|c|c}
    Cases & Max reward & Avg reward & Final $\sigma$ \\
    \hline
TS&-923.397&-1193.720&197.253\\
ES&-258.446&-619.456&321.676\\
PSA&-1107.145&-1938.009&608.504\\
PESA&-2363.198&-3317.970&538.421\\
\textcolor{red}{PPO}&\textcolor{red}{-74.555}&\textcolor{red}{-169.251}&\textcolor{red}{56.243}\\
     \end{tabular}
     \end{minipage}
    \label{tab:85quartercasestats}
\end{table}


\begin{table}[H]
    
    \centering
    \caption{Nemenyi Tests on the best LPs found: 85-quarter case. For the Friedman Test, the p is equal to 0.026 and  0.000 for the Max obj and reward, respectively. Highlighted in red are the successful tests with significance level $\alpha = 0.1$. Most tests for PPO succeeded.}
     \begin{minipage}[b]{0.45\linewidth}
    \begin{tabular}{c|c|c|c|c|c}
    Cases & TS & ES & PSA & PESA & PPO \\
    \hline
TS&1.000&0.351&0.900&0.114&0.114\\
ES&0.351&1.000&0.210&0.900&0.900\\
PSA&0.900&0.210&1.000&\textcolor{red}{0.055}&\textcolor{red}{0.055}\\
PESA&0.114&0.900&\textcolor{red}{0.055}&1.000&0.900\\
PPO&0.114&0.900&\textcolor{red}{0.055}&0.900&1.000\\
    \end{tabular}
    \end{minipage}
    \hspace{15pt}
     \begin{minipage}[b]{0.45\linewidth}
     \begin{tabular}{c|c|c|c|c}
      TS & ES & PSA & PESA &  PPO \\
    \hline
1.000&0.602&0.763&\textcolor{red}{0.081}&\textcolor{red}{0.024}\\
0.602&1.000&\textcolor{red}{0.081}&\textcolor{red}{0.001}&0.522\\
0.763&\textcolor{red}{0.081}&1.000&0.602&\textcolor{red}{0.001}\\
\textcolor{red}{0.081}&\textcolor{red}{0.001}&0.602&1.000&\textcolor{red}{0.001}\\
\textcolor{red}{0.024}&0.522&\textcolor{red}{0.001}&\textcolor{red}{0.001}&1.000\\
     \end{tabular}
     \end{minipage}
    \label{tab:nemenyitests85quartercase}
\end{table}

\paragraph{81-quarter scenario}:
In this last case, which is seemingly harder than the others as explained in Section \ref{sec:extendeddesigns}, ES surpasses PPO in the best pattern found (see Tables \ref{tab:81quartercasestats}). Moreover, PPO exhibits a higher variance in reward (blue curve in Figure \ref{fig:81quartercases}.(5) in appendix \ref{appendix:evolutionoftheaverage}) compared to the other scenarios, which may be due to the complexity of the problem. On the other hand ES, with its global search capacity, sees more low-quality patterns (blue curve in Figure \ref{fig:81quartercases}.(2) is higher than the one in \ref{fig:81quartercases}.(5) in appendix \ref{appendix:evolutionoftheaverage}) but could found a better one overall (red curve in Figure \ref{fig:81quartercases}.(2) against  \ref{fig:81quartercases}.(5) for PPO in appendix \ref{appendix:evolutionoftheaverage}). Additionally, all the NTs involving PPO failed for the first time for the best LPs found (see Table \ref{tab:nemenyitests81quartercase}), while ES succeeded against all but PPO. In spite of exacerbating lower average reward, the large steps of ES allows it to find better solutions than PPO in average. Again, TS surpasses its local search counterpart PSA and PESA is the third best algorithm behind PPO with an average best pattern at -43.585 with a $\sigma$ of 25.374.

\begin{table}[H]
    \centering
    \caption{Best and average objectives, and standard deviation $\sigma$ over the experiments for the best pattern (left) and average pattern in one generation (right)  at the last generation. 81-quarter case}
     \begin{minipage}[b]{0.45\linewidth}
    \begin{tabular}{c|c|c}
   Cases & Avg Max & Final $\sigma$ \\
 \hline

TS & -90.667 & 45.093\\
\textcolor{red}{ES} & \textcolor{red}{-15.230} & \textcolor{red}{6.844}\\
PSA & -139.578 & 167.552\\
PESA &-43.585& 25.374\\
PPO &-32.039& 12.271 \\

    \end{tabular}
    \end{minipage}
     \begin{minipage}[b]{0.45\linewidth}
     \begin{tabular}{c|c|c|c}
    Cases & Max reward & Avg reward & Final $\sigma$ \\
    \hline
TS & -621.317& -1459.519 & 500.883 \\
ES &  -127.881 & -470.282 & 225.089 \\
PSA & -499.030 &-1678.999 & 860.418 \\
PESA & -2298.023 & -3999.596 & 875.587 \\
\textcolor{red}{PPO}& \textcolor{red}{-38.865}& \textcolor{red}{-105.583} & \textcolor{red}{57.831} \\
     \end{tabular}
     \end{minipage}
    \label{tab:81quartercasestats}
\end{table}


\begin{table}[H]
    
    \centering
    \caption{Nemenyi Tests on the best LPs found: 81-quarter case. For the Friedman Test, the p is equal to 0.000 and 0.000 for the Max obj and reward, respectively. Highlighted in red are the successful tests with significance level $\alpha = 0.1$. Most tests for PPO succeeded.}
     \begin{minipage}[b]{0.45\linewidth}
    \begin{tabular}{c|c|c|c|c|c}
    Cases & TS & ES & PSA & PESA & PPO \\
    \hline\
TS&1.000 &\textcolor{red}{0.001}&0.900&0.437&0.210\\
ES&\textcolor{red}{0.001} &1.000&\textcolor{red}{0.001}&\textcolor{red}{0.081}&0.210\\
PSA&0.900 &\textcolor{red}{0.001}&1.000&0.275&0.114\\
PESA&0.437 &\textcolor{red}{0.081}&0.275&1.000&0.900\\
PPO&0.210 &0.210&0.114&0.900&1.000\\
    \end{tabular}
    \end{minipage}
    \hspace{15pt}
     \begin{minipage}[b]{0.45\linewidth}
     \begin{tabular}{c|c|c|c|c}
      TS & ES & PSA & PESA &  PPO \\
    \hline
1.000& 0.210&0.900&0.275&\textcolor{red}{0.003}\\
0.210& 1.000&0.157&\textcolor{red}{0.001}&0.602\\
0.900& 0.157&1.000&0.351&\textcolor{red}{0.002}\\
0.275& \textcolor{red}{0.001}&0.351&1.000&\textcolor{red}{0.001}\\
\textcolor{red}{0.003}& 0.602&\textcolor{red}{0.002}&\textcolor{red}{0.001}&1.000\\
     \end{tabular}
     \end{minipage}
    \label{tab:nemenyitests81quartercase}
\end{table}

\subsection{PPO versus legacy algorithms on (single-objective optimization) with longer runtimes: 89-eighth scenario}
\label{sec:ppoversusbestlong}
In this section we are comparing the algorithms after we have ran 50,000 samples instead of 20,000 and experimented changing some hyper-parameters to see whether we could improve the best LP found by each algorithm for 89-eighth scenario. 
We observed that overall the trends observed in this section are similar to those in the last section, across the other 4 scenarios. Therefore, we left the study of the other 4 scenarios in appendix \ref{appendix:ppoversuslegacyalgorithmsontheextendeddesigns} and detail only one here. 

Additionally, based on our seminal work \cite{seurin2024assessment}, we knew that initializing each episode with the best solution ever found would improve PPO's quality but in preparing this paper we have also observed that increasing \code{n\_steps} to 8 is beneficial for the intricate scenarios. Hence we have decided to set those hyper-parameters for the following studies. 

\begin{figure}[H]
    \centering
    \includegraphics[scale=0.6]{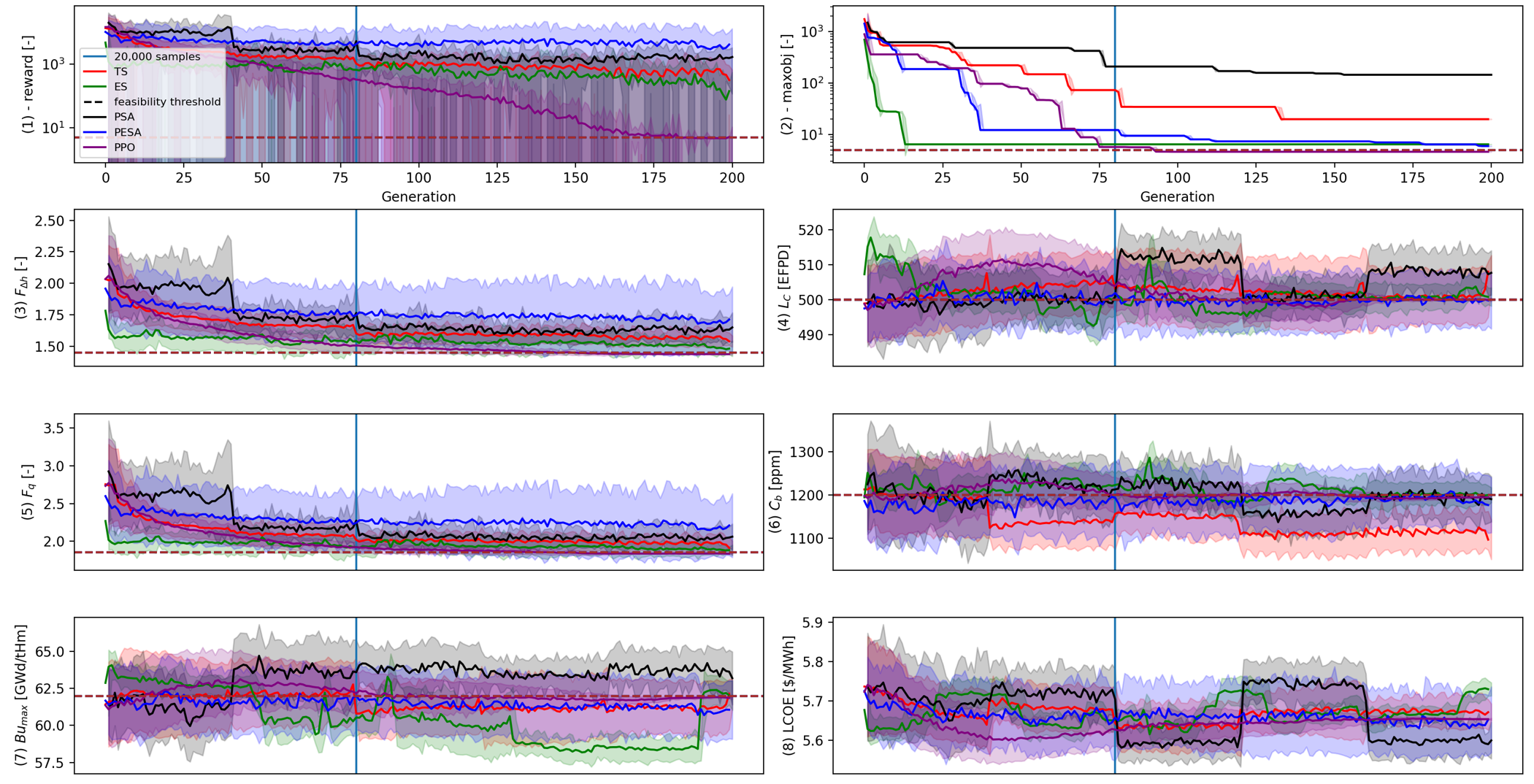}
    \caption{Evolution of the best objective, average reward, and constraints for each algorithm up to 50,000 samples averaged over 200 generations to help with visualization (i.e., one generation contains approximately 250 samples) for the 89-eighth scenario. We have added a vertical line that corresponds to the limit where we stopped collecting samples for the experiments of Section \ref{sec:ppoversuslegacyextended}.}
    \label{fig:89eighthfull}
\end{figure}

\paragraph{89-eighth scenario}: Figure  \ref{fig:89eighthfull} showcases the evolution of the FOMs for each algorithm over the longer optimization runs. The evolution up to 50,000 samples confirms the results obtained in the previous sections. The local search method, TS, continues to improve as observed previously but could not catch up with PESA, ES, or PPO, while PSA got stuck in a local optimum. ES did not show further improvement, whereas PESA managed to catch up, most likely due to its capacity to escape local optimality \cite{radaideh2022pesa}. Nevertheless, PPO remains the best algorithm, with steady improvement thanks to its ability to satisfy the $F_q$ and$F_{\Delta h}$ constraints. Additionally, PPO consistently generates feasible solutions from around generation 170 (approximately 42,500 samples), as illustrated in Figure \ref{fig:89eighthfull}.(1). The best solutions found out of all the experiments by each algorithm are listed in Table \ref{tab:bestfoms89eighthcaselong}, which confirms PPO’s superiority, being the only algorithm that found a feasible solution with an objective value (Obj) greater than or equal to -5.00.

Section \ref{sec:summaryofstudies} summarizes the results obtained from both the statistical studies of Section \ref{sec:ppoversuslegacyextended} and the studies with longer runtimes of Sections \ref{sec:ppoversusbestlong} and \ref{appendix:ppoversuslegacyalgorithmsontheextendeddesigns}.

\begin{table}[H]
    \centering
    \begin{tabular}{c|c|c|c|c|c|c|c}
    \hline
 Cases & Obj & Fq & Fh & Cb & Max Bu & Cy & LCOE\\
 \hline
ES & -6.46&1.844&1.443& 1161.9	& \underline{\textbf{62.311}} & \underline{\textbf{501.3}} & 5.663		\\
TS & -18.110 & \underline{\textbf{1.869}} & \underline{\textbf{1.472}} & 1144.80 & 61.774& \underline{\textbf{506.400}} & 5.617 \\
SA &  -5.920 & 1.821 & \underline{\textbf{1.455}} & 1194.500 & 61.029 & \underline{\textbf{500.600}} & 5.588 \\
PESA & -5.98&1.848 &1.45& \underline{\textbf{1204.4}}& 61.422& \underline{\textbf{500.6}} & 5.608	\\
PPO& \textcolor{red}{-4.605}$^{\star \star}$ & \textcolor{red}{1.815} & \textcolor{red}{1.436} & \textcolor{red}{1178} & \textcolor{red}{61.415} & \textcolor{red}{500} & \textcolor{red}{5.605} \\
    \end{tabular}
    \caption{Best solution found in the 89-eight scenario. Highlighted in red is the best case (PPO). The violated constraints are underlined and highlighted in bold.}
     \label{tab:bestfoms89eighthcaselong}
      \begin{tablenotes}
      \item$^{\star}$ PPO with 'best' as initialization was not ran during statistical analysis, hence this box is not filled.
      \item$^{\star \star}$ An objective Obj greater than -5.00 correspond to a feasible pattern.
    \end{tablenotes}
\end{table}

\subsection{Summary of the studies conducted}
\label{sec:summaryofstudies}

\subsubsection{Summary of the statistical studied conducted}
\label{sec:summaryofthestatisticalstudy}
Table \ref{tab:summaryofstatistics} summarizes the results observed when comparing PPO, PESA, and the legacy SO against each other with statistical tools. For all but the 81-quarter case (see last row), PPO is the winning algorithm. Thanks to its ability to find high-quality solutions fast and the low NTs, PPO can be said to be statistically significantly better for the eighth-cases (see rightmost column of Table \ref{tab:summaryofstatistics}). 
\begin{table}[H]
    \centering
    \begin{tabular}{c|c|c|c|c|c|c|c}
    \hline 
         &  TS & PSA & ES & PESA & PPO & Best (avg) & Best (stats)\\
     \hline
       89-eighth  & poor$^{\star}$ & medium & medium & medium & good & \textcolor{red}{PPO} & \textcolor{red}{True}\\
       81-eighth  &  poor & poor & good & medium & good & \textcolor{red}{PPO} & \textcolor{red}{True}\\
       89-quarter  & poor & poor & medium & medium & good & PPO & False\\
       85-quarter  & medium & poor & good & good & good & PPO & False\\
       81-quarter  & poor & poor & good & medium & good & ES & False\\
    \end{tabular}
    \caption{Summary of the average performance of each algorithm on each problem. The two rightmost column "Best (avg)" and "Best (stats)" mentioned here, relate which algorithm was the best in terms of average performance and if it can be said statistically significantly better, respectively. Highlighted in red are the cases where the latter is true.}
    \label{tab:summaryofstatistics}
    \begin{tablenotes}
    \item $^{\star}$ There are three level of performance: poor, medium, and good. They are arbitrarily based on the overall performance of the best performing algorithms across all problems.
    \end{tablenotes}    
\end{table}
Additional observations can be summarized as follows:
\begin{enumerate}
    \item 
The LP scenarios derived for this work are more amenable to a global search at the beginning to promote higher exploration and find promising directions. Once a set of promising directions is identified, the search must be reduced to a local search to exploit these directions heavily. The trainable weights of the policy in PPO can adapt to function as both a global search initially and a local search subsequently.
\item
PPO performs very well across all problems studied in Section \ref{sec:ppoversuslegacyextended}. We also used similar hyperparameters in \cite{seurin2024assessment} for classical benchmarks of the CEC17 test suite \cite{awad2017problem} and demonstrated PPO’s good performance on those as well. These studies confirm that PPO’s ability to dynamically learn a policy allows it to adapt its search based on the varying curvature of objective spaces. As a result, PPO has the potential to perform well across many different problems.
\item
This consistent improvement is also observed in Figures \ref{fig:89eighthcases} through \ref{fig:81quartercases}, where both the reward and the maximum objective consistently increase, indicating that PPO keeps improving over time. While TS shows some improvement, the speed at which PPO progresses is much higher, consistently surpassing TS.
\item
Contrary to our expectations, utilizing PESA, which combines ES and PSA, did not improve upon ES, with its performance falling between ES and TS. We hypothesize that an ablation study must be conducted to efficiently incorporate all hyperparameters within PESA to surpass ES or to increase the number of available processors. Nevertheless, it could also improves ES and PPO's performances.
\end{enumerate}

\subsubsection{Summary of the longer studies conducted}
\label{sec:summary of the studies conducted}

Table  \ref{tab:summaryofstatisticslong} summarizes how each algorithm improves beyond the 20,000 samples mark for each scenario. Overall, TS and PESA are the algorithms that improved the most with a larger number of samples. However, this improvement is not sufficient to surpass ES and PPO, with PPO showing excellent performance very quickly. On the other hand, PSA barely improved and got stuck in bad local optima in all scenarios.

\begin{table}[H]
    \centering
    \begin{tabular}{c|c|c|c|c|c|c}
    \hline 
        Scenarios &   TS & PSA & ES & PESA & PPO &  Best (avg)\\
     \hline
       89-eighth  & strong$^{\star}$ & poor & marginal & strong & marginal & \textcolor{red}{PPO}\\
       81-eighth  &  poor & poor & strong & poor  & strong &\textcolor{red}{PPO}\\
       89-quarter  & strong & marginal & marginal & marginal & marginal &  PPO\\
       85-quarter  & marginal & marginal & marginal & strong & marginal &  PPO\\
       81-quarter  & strong &  poor & strong &  strong & marginal & PESA \\
    \end{tabular}
    \caption{Summary of the improvement in performance of each algorithm on each problem for the longer run-times. PPO remains the best algorithms for all but the 81-quarter scenario}
    \label{tab:summaryofstatisticslong}
    \begin{tablenotes}
        \item $^{\star}$ There are 3 levels of improvement: (1) poor if limited or no improvement was observed by running up to 50,000 samples, (2) marginal if some improvements were observed but did not result in drastically better performance, and (3) strong if the improvement resulted in noticeable benefit for the final loading pattern found (e.g., PPO managed to find a feasible pattern with 81-eighth well after collecting 20,000 samples).
    \end{tablenotes}
\end{table}
Additional observations can be summarized as follows:
\begin{enumerate}
    \item Contrary to our hypothesis in Section \ref{sec:summaryofthestatisticalstudy}  regarding local search methods (PSA and TS), which suggested they would steadily improve and potentially catch up with global search methods like ES or PPO after 50,000 samples, they failed and often got stuck in local optima (e.g., in the 81-eighth scenario). As a result, they never caught up in our 5 scenarios. However, PESA managed to catch up with ES or PPO in 3 out of 5 scenarios (all except the 81-eighth and quarter scenarios).
    \item Increasing the number of samples exacerbated most of the results observed in the previous Section  \ref{sec:ppoversuslegacyextended}: PPO still outperforms and sometimes converges to only generating feasible solutions (e.g., in the 89-quarter scenario), while ES improved but remained worse than PPO except for the 81-quarter case.
    \item PPO could find a feasible solution for 3 out of 5 scenarios. For the 85-quarter scenario, it was very close, with only $L_{cy}$ equal to 500.8 instead of 500.0. With our classical constraints of 500.0 $\le L_{cy}$ \cite{seurin2024assessment,seurin2024multiobjective,seurin2024physics}, it would have been feasible, while ES could only find one even with the relaxed constraints, further demonstrating PPO’s superiority.
    \item TS can sometimes significantly outperform its local search counterpart PSA, which almost never improved even after 20,000 samples, as emphasized in Table \ref{tab:summaryofstatisticslong} after running the optimization up to 50,000 samples (e.g., in the 89-quarter scenario). However, TS is always worse than PPO, ES, and PESA.
    \item Based on the problems described in this work, we rank the algorithms as follows: PPO, ES, PESA, TS, and PSA. While ES and PESA found good solutions early, they failed to improve sufficiently over the longer run to surpass PPO (except for the 81-quarter scenario). Local search methods such as TS improve steadily but too slowly and could not outperform PPO at 50,000 samples, which is prohibitive, since collecting that many samples can take more than two days (e.g., the quarter scenarios). Table  \ref{tab:rankorderandexplanation} summarizes the explanations we have given for this ordering:
\begin{table}[H]
    \centering
    \begin{tabular}{c|c|c|c}
    \hline
   Algorithms   &  Type & Rank order  &  Explanation\\
   \hline
    PPO     & 'Both'$^{\star}$ adaptive & 1 & \makecell{largely (statistically) superior \\ on most benchmarks (4 out of 5)}\\
   \hline
    ES & 'Global' & 2 & \makecell{consistently good but fail to find \\ feasible solutions on \\ most benchmarks unlike PPO \\ (e.g., 89/81-eighth scenarios)}\\
    \hline
    PESA & 'Both' ensemble & 3 &\makecell{not consistent and less sample efficient\\ but can equate PPO and ES \\ in some scenarios (e.g., 89/85-quarter)}\\
     \hline
    TS & 'Local' & 4 & \makecell{poor performance except for \\ the 89-quarter scenario. Can better in performance \\ if larger number of samples ran (but prohibitive)}\\
     \hline
    PSA & 'Local' & 5 & \makecell{poor performance on all benchmarks \\ with early convergence to local optima \\ in most scenarios}\\
    \end{tabular}
    \caption{Algorithms tested in this work, type, Ranking,  and rationale behind the ordering for the problem scenarios described in Section \ref{sec:testsets}.}
    \label{tab:rankorderandexplanation}
    \begin{tablenotes}
    \item $^{\star}$ There are three possible types; (1) 'Global' search, (2) 'Local' search, and (3) 'Both' a global and local search. For type 'Both', we have adaptive methods like PPO, and ensemble methods like PESA.
\end{tablenotes}
\end{table}
\end{enumerate}

\section{Concluding remarks}
\label{sec:concludingremarks}

In this work, we have demonstrated the statistical superiority of leveraging PPO, a recently developed deep learning-based algorithm, over classical methods for PWR LP optimization (see Section \ref{sec:ppoversuslegacyextended}). Moreover, running the optimization longer in Section \ref{sec:ppoversusbestlong} not only confirmed the statistical analysis performed on shorter runs but also exacerbated the differences among most algorithms: only PESA and TS showed significant improvement, while PPO remained the best on 4 out of 5 benchmarks. When PPO finds high-quality solutions, it also increases the certainty of finding them, which is crucial for its acceptance and deployment at NPPs. Therefore, PPO is clearly the best algorithm for the defined scenarios.

We attribute the superior performance of PPO against legacy algorithms to its adaptive policy $\pi_{\theta}$ with learnable parameters $\theta$: PPO initially behaves as a global search with a highly stochastic $\pi_{\theta}$,  and then as a local search once promising directions in the objective space are identified through intelligent refinement of the weights $\theta$. Therefore, we termed our approach of using PPO as \textit{automatic supervised setting} in past research \cite{seurin2024multiobjective}, but more specifically as 'Both' adaptive in Table \ref{tab:rankorderandexplanation} in this research. This adaptive behavior combines high quality, stability, and fast improvement (also known as Sample Efficiency (SE) \cite{seurin2022pwr,seurin2024assessment}). 

Nevertheless, in the set of 5 scenarios studied (i.e., 89-eighth, 81-eighth, 89-quarter, 85-quarter, and 81-quarter), we did not assume a good solution was known in advance. If that were the case, we could easily initialize the legacy methods with it (e.g., input the known solution in the initial population in ES or start the first Markov Chains of PSA), which might enable early improvement of the solution. However, such initialization is not possible with PPO; we would need to first train PPO to recover that solution before improving it. This is a classical approach for solving combinatorial optimization problems with RL in the computational mathematics literature \cite{seurin2024assessment,mazyavkina2021reinforcement}. Even though literature suggests that PPO surpasses classical heuristics in these instances too, training PPO starting from a known good solution and comparing it against legacy methods in the context of LP optimization is an area of ongoing research.

\section*{Credit Authorship Contribution Statement}

 \textbf{Paul Seurin:} Conceptualization, Methodology, Software, formal analysis \& investigation, Writing– original draft preparation, Visualization. \textbf{Koroush Shirvan:} Conceptualization, investigation, Writing– review \& editing, Supervision, Funding acquisition. All authors have read and agreed to the published version of the  manuscript.
 
 \section*{Declaration of Competing Interest}
 The authors declare that they have no known competing financial interests or personal relationships that could have appeared
 to influence the work reported in this paper.

\section*{Acknowledgement}
This work is sponsored by Constellation (formerly known as Exelon) Corporation under the award (40008739).

\normalsize
\appendix
\section{Hyper-parameter selection for PSA and ES}
\label{appendix:hyperparameterforpsaandes}
In this section, in order to not bias our successful results towards the RL's MDP and action spaces from \cite{seurin2024assessment}, we attempted to find optimum hyper-parameters for PSA and ES for the input space defined in \ref{sec:inputdecision}. Nonetheless, we omit the study for TS as using\code{ncores} equal to 32, \code{rank} for \code{reinforce\_best}, \code{m} equal to 5, and scale \code{chain\_size} to 10 to obtain at most similar number of samples resulted in good performances. We utilize 32 cores for SO-based methods, which amount to 32 Markov Chains in PSA and 32 individuals in the population for ES. On top of that, the problem defined in \cite{seurin2022pwr,seurin2024assessment} requires a high level of exploitation. Therefore, we focus our search for the SO-based algorithms to hyper-parameter values which imply high exploitation. For PSA we focus on \code{$\chi$} and \code{Tmin}. The former was explained in section \ref{sec:simulatedannealing}, while the latter controls the acceptance probability. Higher \code{$\chi$} may induce higher exploration by performing bigger jumps in the objective space. Lower \code{Tmin} results in higher exploration by accepting more solutions. For ES we focus on \code{$\mu$} and \code{mutpb}. The former was explained in section \ref{sec:geneticalgorithm} while the latter is the probability for an individual to undergo a mutation. Higher values for both induce more exploration.

Tables \ref{tab:psahyperparameters} and \ref{tab:eshyperparameters} showcase the best objectives found for both PSA and ES, the number of samples before convergence for PSA, and the mean objective SE (defined in \cite{seurin2022pwr,seurin2024assessment} as the average reward at the last generation) for ES. Indeed, the criteria for converging for PSA is based on the average number of accepted solutions per chain during the \code{chain\_size} operations. An early convergence may reduce the probability to find a better pattern by not traversing the search space exhaustively.

In light of table \ref{tab:psahyperparameters}, we can see that there is a trade-off to find for \code{$\chi$}. Lower value takes longer to converge and do not improve quickly enough. The case \code{$\chi$} equal to 0.1 found a good pattern but converged very early. Reducing \code{Tmin} to 0.005 helped recover a good pattern while increasing the number of steps. For this reason, we decided to pursue the latter case as more steps may induce a higher chance to find a better optimum.

In light of table \ref{tab:eshyperparameters}, it is clear that case 2 + \code{mutpb} = 0.3 outperform the others. It has a high best reward while having a low mean objective (SE). Therefore, it still has a large potential for exploration compared to the others while already finding the best loading pattern by far.
The values for these hyper-parameters will be kept to compare RL (in particular PPO) against SO-based approaches in section \ref{sec:ppoversuslegacyextended}.
\begin{table}[H]
    \centering
    \caption{Study of the performance of PSA with different \code{$\chi$} values. Except for the last two last rows \code{Tmin} is set at 1.}
    \begin{tabular}{c|c|c}
    \hline
   \code{$\chi$} &  Best Objective & Samples before convergence\\
    \hline
      0.2  & -74.811 & 7531\\
       0.1  & -10.156   & 7531\\
       0.05 &    -300.455     & \textcolor{red}{36000} \\
       0.1 + \code{Tmin} = 0.05 &  -80.544  & 9633\\
       0.1 + \code{Tmin} = 0.005 & \textcolor{red}{-15.868} & 15777 \\
       \hline
    \end{tabular}
    \label{tab:psahyperparameters}
\end{table}

\begin{table}[H]
    \centering
   \caption{Study of the performance of ES with different \code{$\mu$} values. With $\mu = 1$, we utilize \code{cxpb} = 0 to cancel crossovers. For the other tests, \code{cxpb} = 0.65 and \code{mutpb} = 0.2 (i.e., the probability that an individual is just copied is 15\%).}
    \begin{tabular}{c|c|c}
    \hline
   \code{$\mu$} &  Best Objective & Mean objective SE \\
    \hline
       30&-207.883 & -16179.675 \\
       15&  -59.753 & -584.244 \\
        5& -56.518 & -118.312 \\
        2&  -29.890  &-336.882 \\
        1&  -50.118    &   -1445.922   \\
        2 + \code{muptb} = 0.1 & -32.816 & -867.490 \\
        2 + \code{mutpb} = 0.3 & \textcolor{red}{-13.632} &  \textcolor{red}{-1986.772} \\
        \hline
    \end{tabular}
    \label{tab:eshyperparameters}
\end{table}

\section{Evolution of the average and max rewards across the experiments for each algorithm}
\label{appendix:evolutionoftheaverage}

\begin{figure}[H]
    \centering
    \includegraphics[scale=0.2]{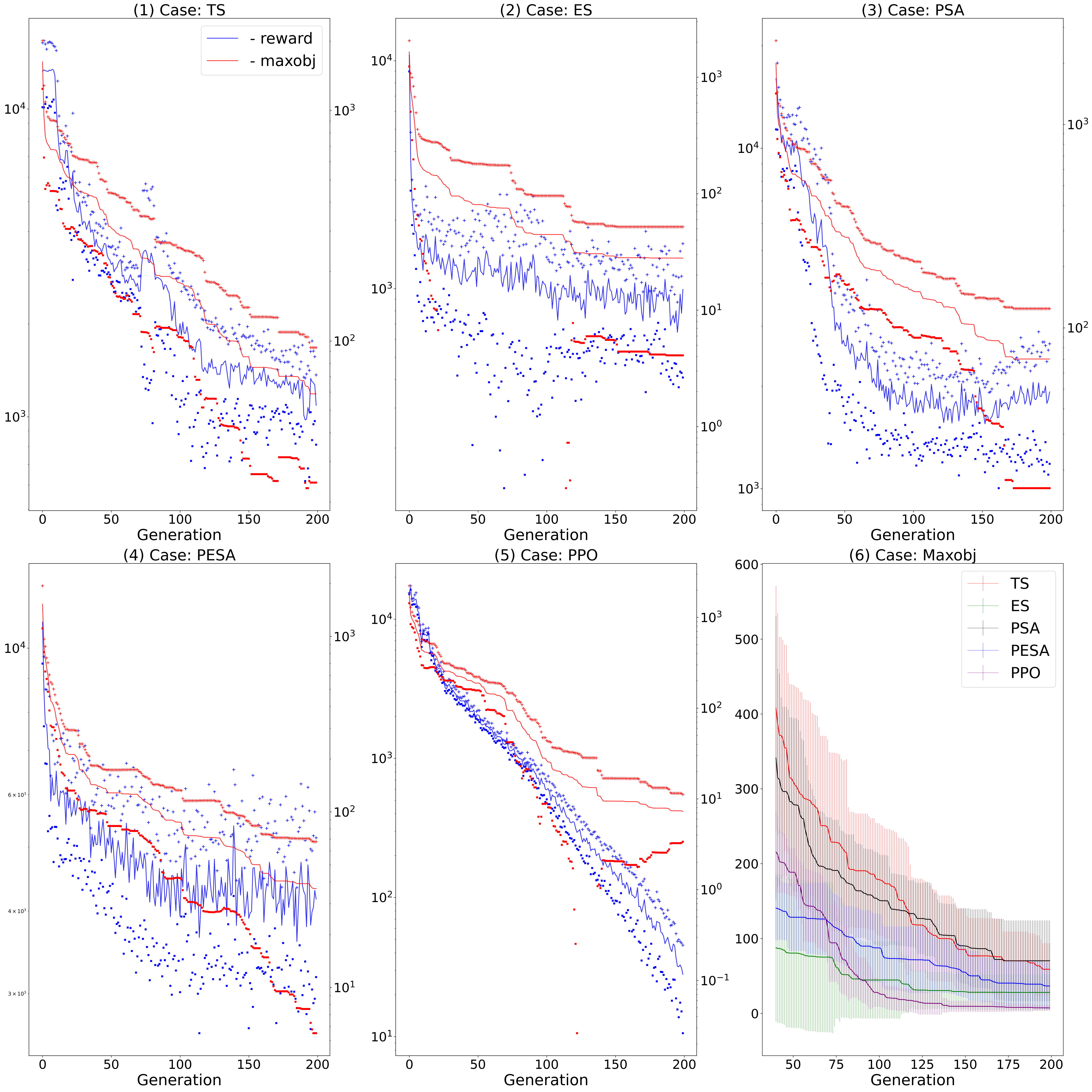}
    \caption{Evolution of the mean score (blue curves) and max objective (red curves) per 200 generation, for the 89-eighth scenario. The scattered elements surrounding the lines are the mean $\mu$ $\pm$ the deviation $\sigma$ over the 10 experiments.}
    \label{fig:89eighthcases}
\end{figure}

\begin{figure}[H]
    \centering
    \includegraphics[scale=0.2]{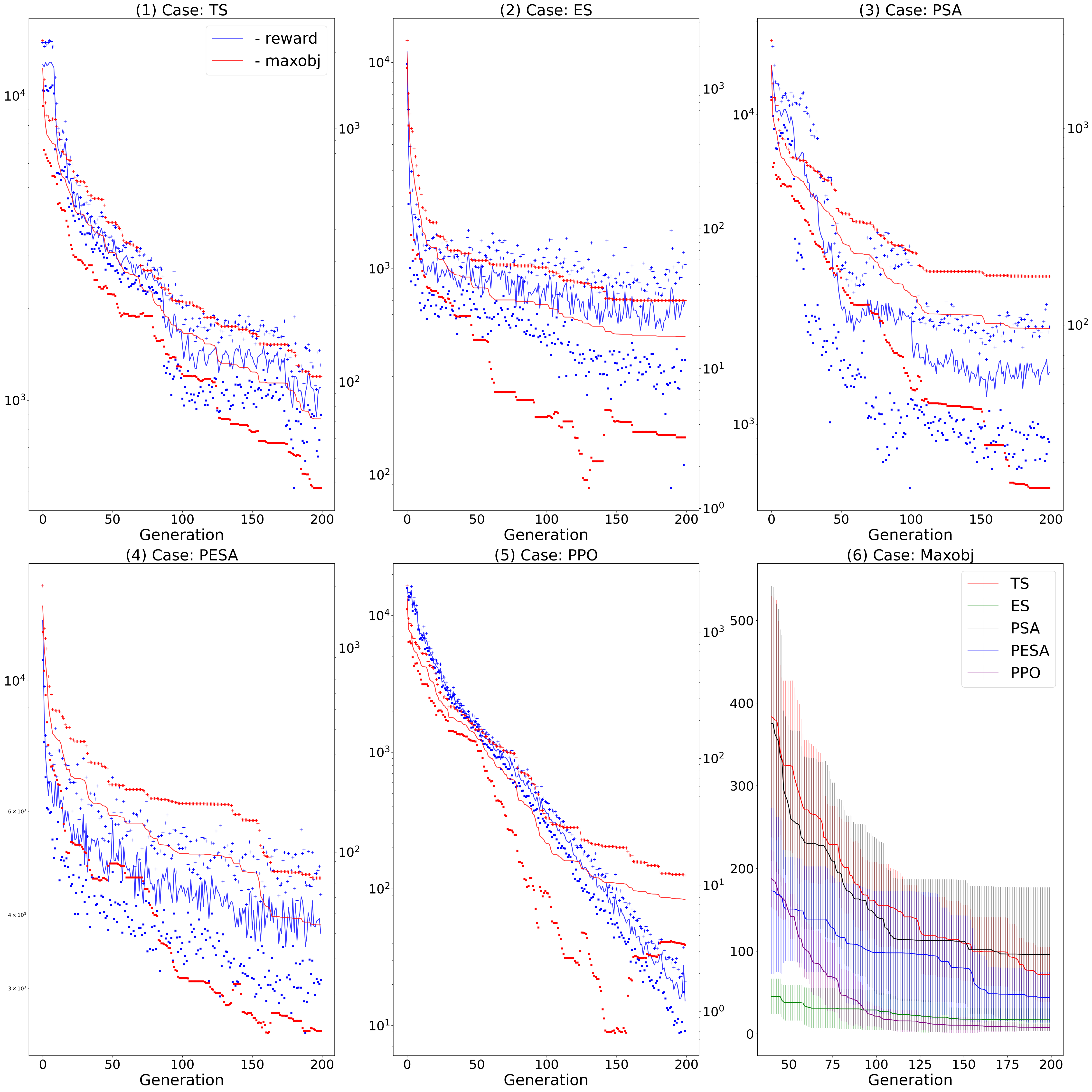}
    \caption{Evolution of the mean score (blue curves) and max objective (red curves) per 200 generation, for the 81-eighth scenario. The scattered elements surrounding the lines are the mean $\mu$ $\pm$ the deviation $\sigma$ over the 10 experiments}
    \label{fig:81eighthcases}
\end{figure}

\begin{figure}[H]
    \centering
    \includegraphics[scale=0.2]{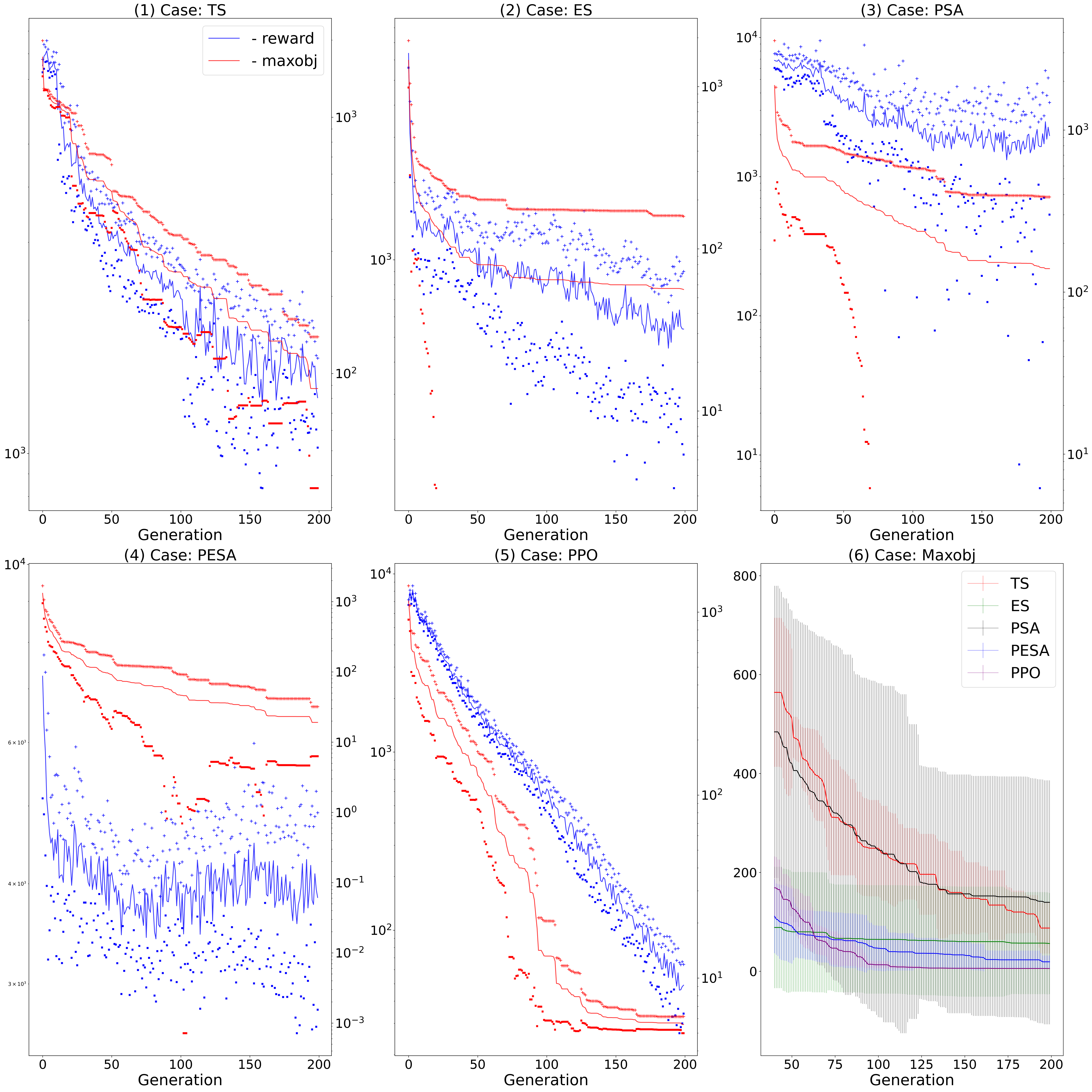}
    \caption{Evolution of the mean score (blue curves) and max objective (red curves) per 200 generation, for the 89-quarter scenario. The scattered elements surrounding the lines are the mean $\mu$ $\pm$ the deviation $\sigma$ over the 10 experiments}
    \label{fig:89quartercases}
\end{figure}

\begin{figure}[H]
    \centering
    \includegraphics[scale=0.2]{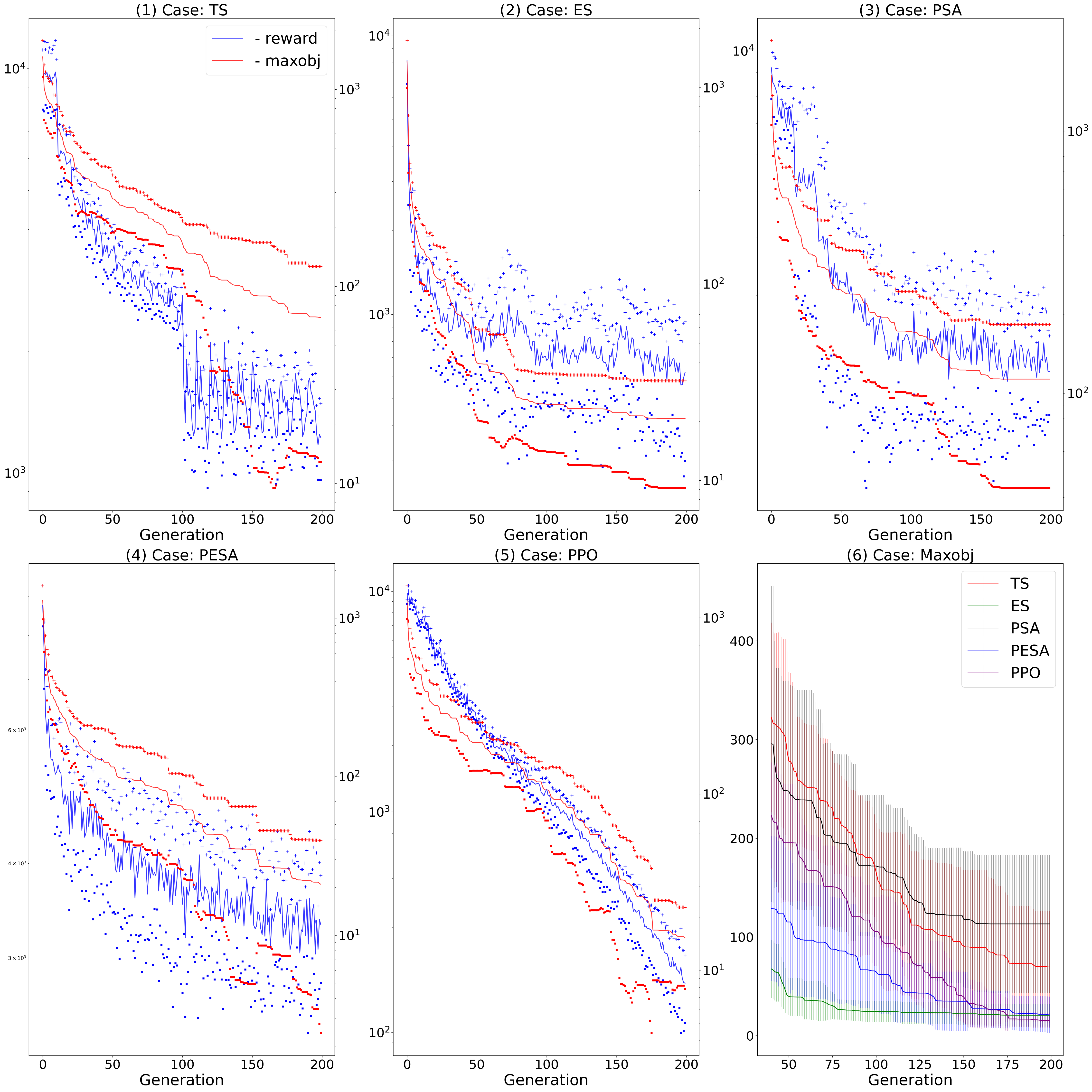}
    \caption{Evolution of the mean score (blue curves) and max objective (red curves) per 200 generation, for the 85-quarter scenario. The scattered elements surrounding the lines are the mean $\mu$ $\pm$ the deviation $\sigma$ over the 10 experiments}
    \label{fig:85quartercases}
\end{figure}

\begin{figure}[H]
    \centering
    \includegraphics[scale=0.2]{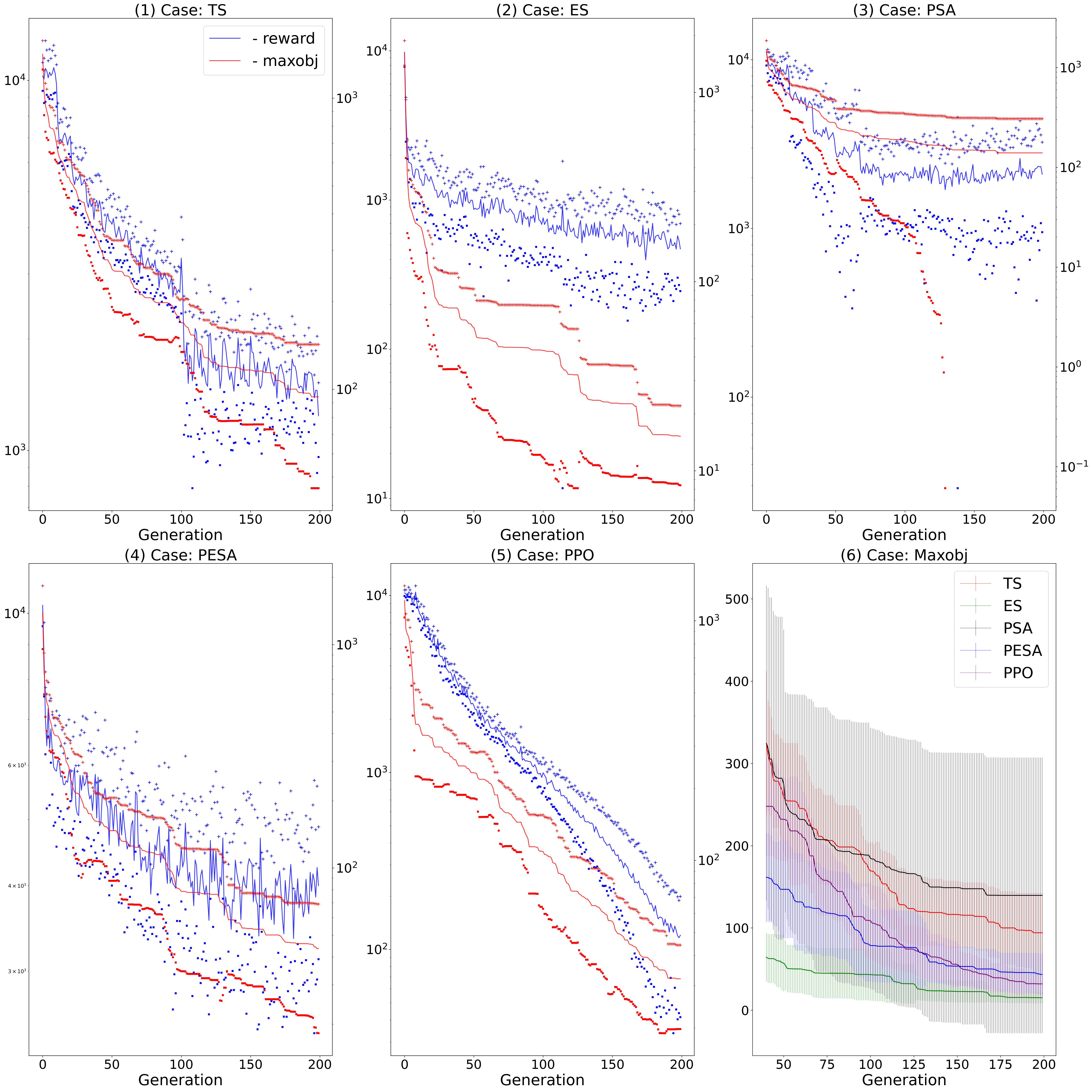}
    \caption{Evolution of the mean score (blue curves) and max objective (red curves) per 200 generation, for the 81-quarter scenario. The scattered elements surrounding the lines are the mean $\mu$ $\pm$ the deviation $\sigma$ over the 10 experiments}
    \label{fig:81quartercases}
\end{figure}

\section{PPO versus legacy algorithms on the extended designs (single-objective optimization) with longer runtimes}
\label{appendix:ppoversuslegacyalgorithmsontheextendeddesigns}
This section presents the remainder of the study in Section \ref{sec:ppoversusbestlong} which studied the behavior of each algorithm over longer runtimes for the 89-eighth scenario. We start with the 81-eighth scenario.

\begin{figure}[H]
    \centering
    \includegraphics[scale=0.6]{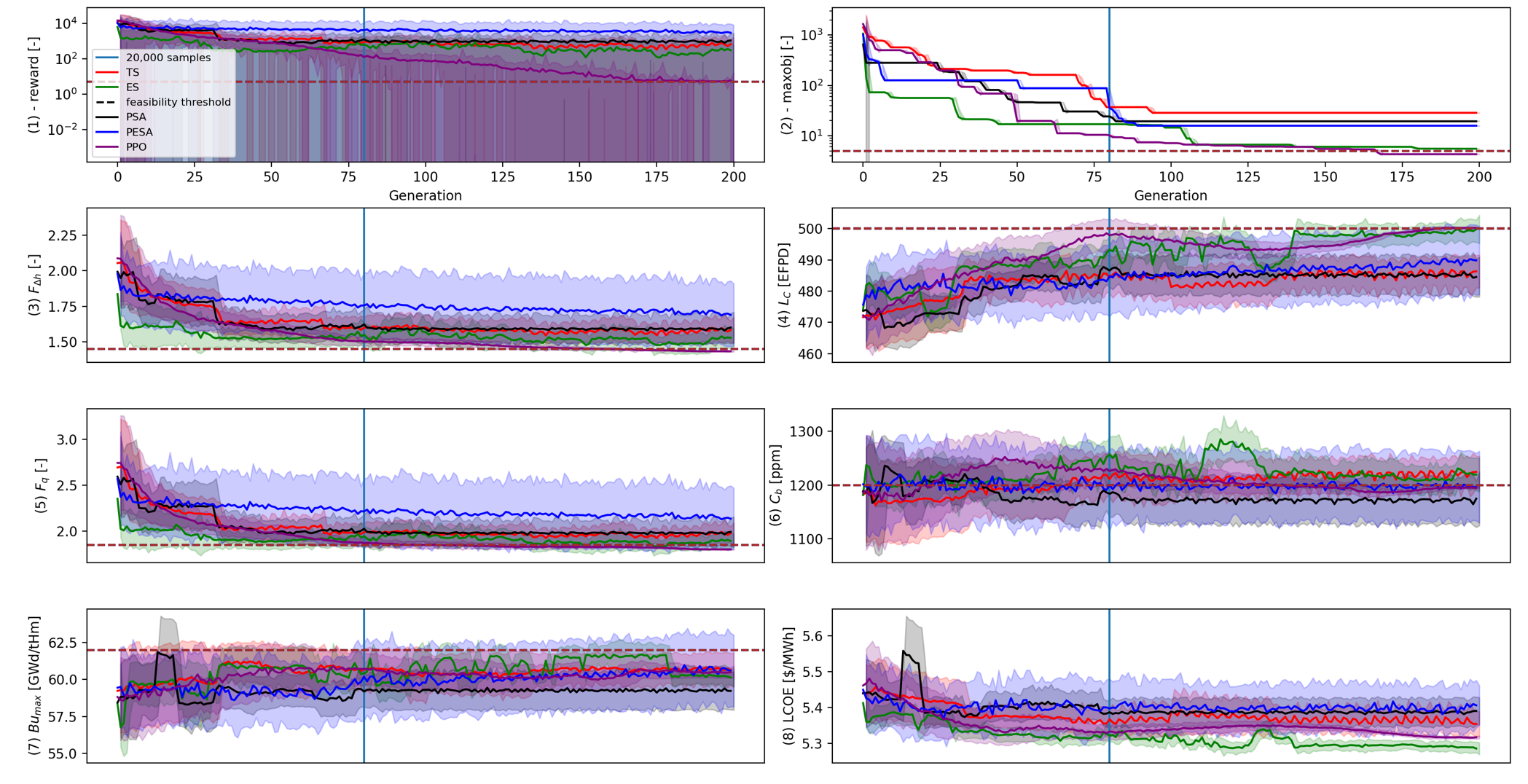}
    \caption{Evolution of the best objective, average reward, and constraints for each algorithm up to 50,000 samples averaged over 200 generations to help with visualization (i.e., one generation contains approximately 250 samples) for the 81-eighth scenario. We have added a vertical line that corresponds to the limit where we stopped collecting samples for the experiments of Section \ref{sec:ppoversuslegacyextended}.}
    \label{fig:81eighthfull}
\end{figure}

\paragraph{81-eighth scenario}: In this more complex scenario, PPO still managed to find a feasible solution, as observed in Figure \ref{fig:81eighthfull}  and Table \ref{tab:bestfoms81eighthcaselong}. Unlike the 89-eighth scenario, ES continuously improved and significantly outperformed PESA, PSA, and TS, all of which got stuck in local optima. However, ES could not outperform PPO. The superiority of PPO and ES is characterized by a high $L_{cy}$ (see Figure  \ref{fig:81eighthfull}..(4)), where both ES and PPO balanced a high $L_{cy}$ and a low $C_b$ (see Figure  \ref{fig:81eighthfull}.(6)), while also achieving good peakings  (see Figures \ref{fig:81eighthfull}.(3) and (5)). This was the challenge of this scenario, as explained in Section \ref{sec:extendeddesigns}. However, PPO is the only algorithm that achieved the constraint of $L_{cy}$ equal to 500, whereas ES could not increase $L_{cy}$ without significantly increasing $C_b$, resulting in a local optimum of $L_{cy}$ and $C_b$ equal to 498.6 EFPD and 1201.5 ppm, respectively. Additionally, PPO consistently generated feasible solutions from around generation 175 (approximately 43,750 samples), as shown in \ref{fig:81eighthfull}.(1).

\begin{table}[H]
    \centering
    \begin{tabular}{c|c|c|c|c|c|c|c}
    \hline
 Cases & Obj & Fq & Fh & Cb & Max Bu & Cy & LCOE \\
 \hline
ES & -5.53 &	1.838 & 	1.447	& \underline{\textbf{1201.5}} &	60.226 & 	\underline{\textbf{498.6}} & 5.294 \\
TS & -28.61	& 	\underline{\textbf{1.865}} &	\underline{\textbf{1.489}} &	\underline{\textbf{1212.4}} &	61.598 & \underline{\textbf{497}} & 5.310  \\
PSA & -14.130 & 1.806 & \underline{\textbf{1.455}} & \underline{\textbf{1207.200}} & 57.466 & \underline{\textbf{491.300}} & 5.367 \\
PESA & -9.520& 1.835 & \underline{\textbf{1.465}} &1199.3 & 61.258 & \underline{\textbf{496.1}}& 5.327 \\
PPO& \textcolor{red}{-4.308} & \textcolor{red}{1.802} &	\textcolor{red}{1.432} &	\textcolor{red}{1198.9} &	\textcolor{red}{60.394} & \textcolor{red}{500}&\textcolor{red}{5.308} \\
    \end{tabular}
    \caption{Best solution found in the 81-eight scenario. Highlighted in red is the best case (PPO). The violated constraints are underlined and  highlighted in bold.}
     \label{tab:bestfoms81eighthcaselong}
\end{table}

\begin{figure}[H]
    \centering
    \includegraphics[scale=0.6]{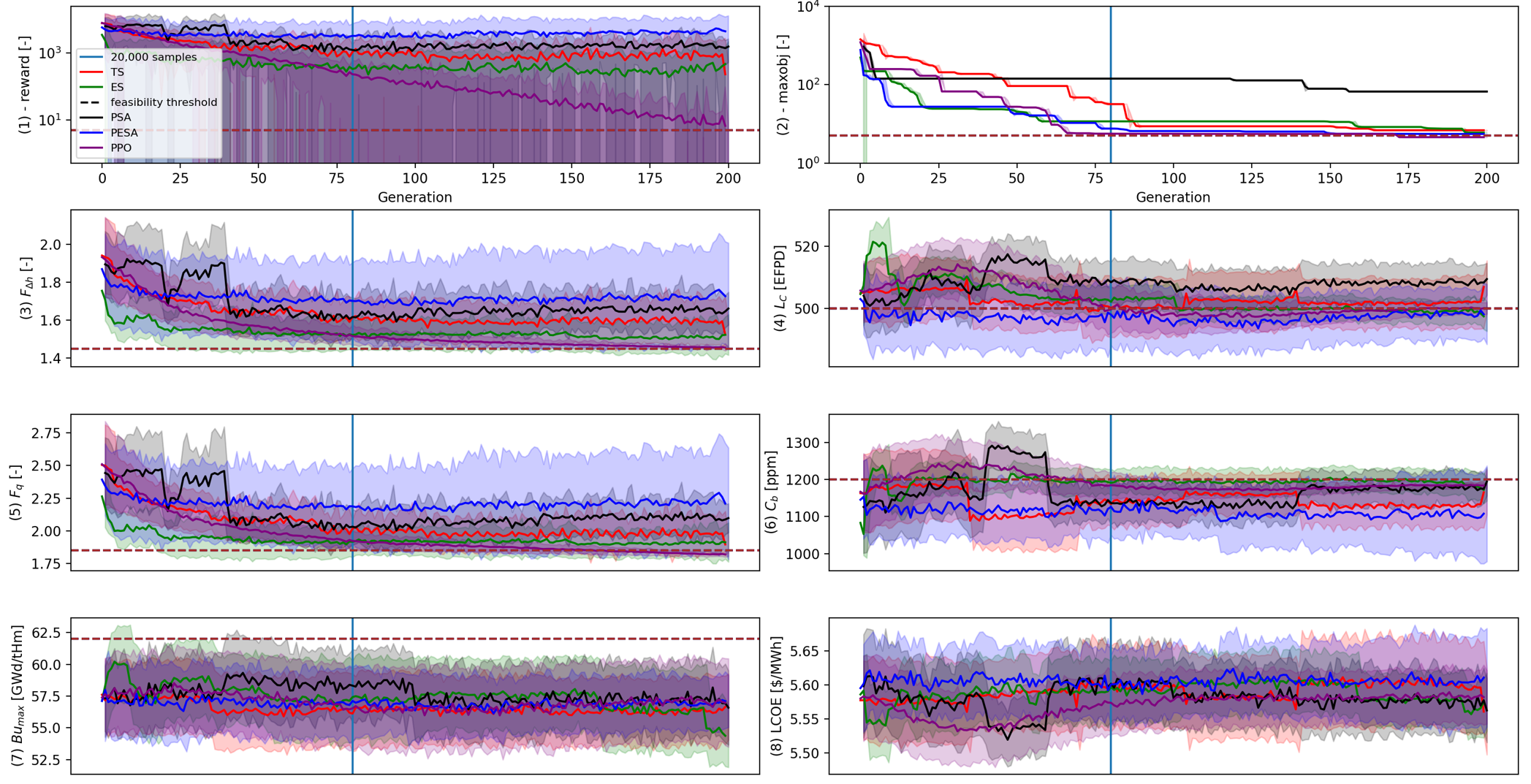}
    \caption{Evolution of the best objective, average reward, and constraints for each algorithm up to 50,000 samples averaged over 200 generations to help with visualization (i.e., one generation contains approximately 250 samples) for the 89-quarter scenario. We have added a vertical line that corresponds to the limit where we stopped collecting samples for the experiments of Section \ref{sec:ppoversuslegacyextended}.}
    \label{fig:89quarterfull}
\end{figure}

\paragraph{89-quarter scenario}: Interestingly, every algorithms managed to find close to feasible solutions as showcased in Table \ref{tab:bestfoms89quartercaselong}. Again, PPO and ES are superior by being the only two algorithms that found a feasible solution, with the PPO one being of lower LCOE (5.516\$/MWh against 5.526 \$/MWh). The close performance of all algorithms might be explained by comparing to the 89-eighth, where the peak pin burnup $Bu_{max}$ was harder to reach, as can be seen on Figures \ref{fig:89quarterfull}.(7) against \ref{fig:89eighthfull}.(7).
Again, PPO under-performed early compared to ES but also PESA and caught up around the 70th generation (about 17,500 samples), and remain better and nearly converged to a feasible pattern at each episode around the end, as seen in Figure \ref{fig:89quarterfull}.(1) and .(2). Here again, TS starts to largely outperforms PSA after the 20,000 samples mark, which performed poorly even with late improvements after the 150th generation during optimization (see the black curve on Figure \ref{fig:89quarterfull}.(2)).

\begin{table}[H]
    \centering
    \begin{tabular}{c|c|c|c|c|c|c|c}
    \hline
 Cases & Obj & Fq & Fh & Cb & Max Bu & Cy & LCOE \\
 \hline
TS & -6.88	& \underline{\textbf{1.856}}	& \underline{\textbf{1.459}} &	1101 &	51.647 & \underline{\textbf{499}} & 5.550	\\
ES & -4.526&1.846&1.446&1166.8&51.319&500&5.526 \\
PSA & -5.611 & 1.844 & \underline{\textbf{1.452}} & 1070.3 & 60.161 & \underline{\textbf{500.2}} & 5.560 \\
PESA & -5.54 &	1.847 &	\underline{\textbf{1.451}} &	1175.6	& 58.387 &\underline{\textbf{500.1}} & 5.524 \\
PPO& \textcolor{red}{-4.516}&\textcolor{red}{1.820}&\textcolor{red}{1.450}&\textcolor{red}{1154.1}&\textcolor{red}{49.794}&\textcolor{red}{500}&\textcolor{red}{5.516} \\
    \end{tabular}
    \caption{Best solution found in the 89-quarter scenario. Highlighted in red is the best case (PPO). The violated constraints are underlined and  highlighted in bold.}
     \label{tab:bestfoms89quartercaselong}
\end{table}

\begin{figure}[H]
    \centering
    \includegraphics[scale=0.6]{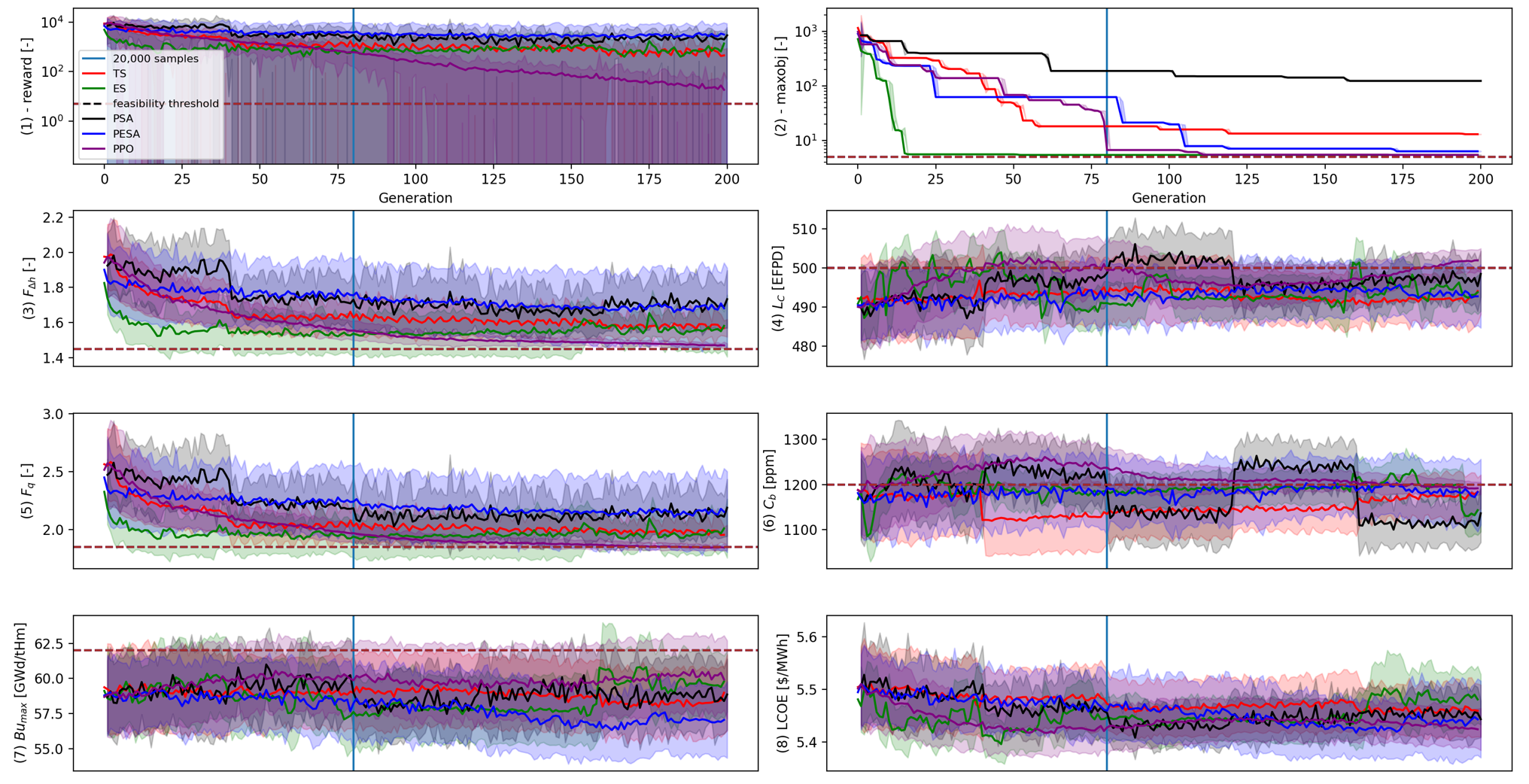}
    \caption{Evolution of the best objective, average reward, and constraints for each algorithm up to 50,000 samples averaged over 200 generations to help with visualization (i.e., one generation contains approximately 250 samples) for the 85-quarter scenario. We have added a vertical line that corresponds to the limit where we stopped collecting samples for the experiments of Section \ref{sec:ppoversuslegacyextended}.}
    \label{fig:85quarterfull}
\end{figure}

\paragraph{85-quarter scenario}: This is the hardest scenario in terms of the number of potential solutions but not the most intricate in terms of striking the right balance between the different FOMs, as introduced in Section \ref{sec:extendeddesigns}. Here again, ES surpasses every algorithm early on and remains the best until PPO surpasses it. PESA also notably caught up with PPO and ES in terms of the best pattern found (see Figure \ref{fig:85quarterfull}.(2)), but still underperforms compared to the latter (see Table \ref{tab:bestfoms85quartercaselong}). TS competed with PPO early on but then only improved very slowly, while PSA exhibited even less improvement. Moreover, no algorithms could find a feasible solution this time, as seen in Table \ref{tab:bestfoms85quartercaselong} where all the objective values (Obj) are below the feasible threshold of -5.00. However, based on the limits in \cite{seurin2024assessment,seurin2024multiobjective,seurin2024physics}, the solution found by PPO would be considered feasible because only the $L_{cy}$ is 500.8 EFPD, not exactly 500, while ES is also close.

\begin{table}[H]
    \centering
    \begin{tabular}{c|c|c|c|c|c|c|c}
    \hline
 Cases & Obj & Fq & Fh & Cb & Max Bu & Cy & LCOE \\
 \hline
TS & -11.768 & 1.849 & \underline{\textbf{1.469}} & 1187.0 & 58.873 & \underline{\textbf{504.5}} & 5.450 \\
ES & -5.440	& 1.828	& \underline{\textbf{1.451}} &	1168.3	& 58.691  &	\underline{\textbf{499.4}} & 5.395 \\
PSA & - 9.210 & \underline{\textbf{1.857}} & \underline{\textbf{1.467}} & 1151.6 & 58.216 & \underline{\textbf{499.4}} & 5.379 \\
PESA & -5.973 & 1.844 & \underline{\textbf{1.456}} & 1109.0 & 60.137 & \underline{\textbf{499.4}} &  5.509 \\
PPO& \textcolor{red}{-5.426} & \textcolor{red}{1.828} & \textcolor{red}{1.449} & \textcolor{red}{1184.0} & \textcolor{red}{53.619} & \textcolor{red}{\underline{\textbf{500.8}}} & \textcolor{red}{5.362} \\
    \end{tabular}
    \caption{Best solution found in the 85-quarter scenario. Highlighted in red is the best case (PPO). The violated constraints are underlined and  highlighted in bold.}
     \label{tab:bestfoms85quartercaselong}
\end{table}

\begin{figure}[H]
    \centering
    \includegraphics[scale=0.6]{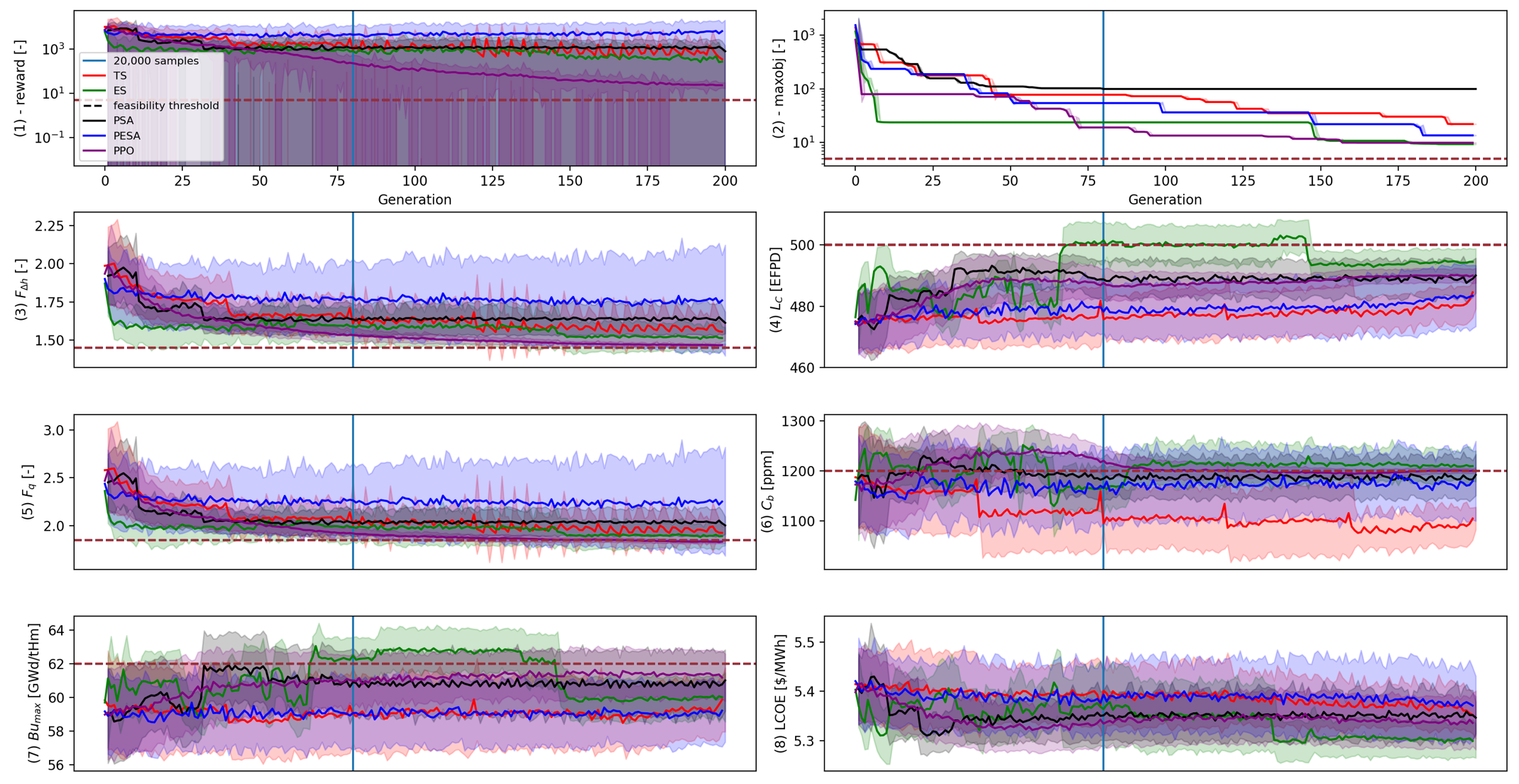}
    \caption{Evolution of the best objective, average reward, and constraints for each algorithm up to 50,000 samples averaged over 200 generations to help with visualization (i.e., one generation contains approximately 250 samples) for the 81-quarter scenario. We have added a vertical line that corresponds to the limit where we stopped collecting samples for the experiments of Section \ref{sec:ppoversuslegacyextended}.}
    \label{fig:81quarterfull}
\end{figure}

\paragraph{81-quarter scenario}: This is the most complex case, and all algorithms struggled. In this scenario, TS managed to keep improving, well beyond PSA’s performance. As observed throughout this section, ES starts to perform well early, but PPO caught up by the limit of 20,000 samples, as seen in Figure \ref{fig:81quarterfull}.(2). However, even though PESA gets close around generation 150, it remains below the performance of ES and PPO. The best pattern found by PPO remains superior to the ES one from generation 75 to 150, but then ES slightly surpasses it.

It is important to note the discrepancy between the seemingly best objectives found in Figure \ref{fig:81quarterfull}.(2) and those showcased in Table  \ref{tab:bestfoms81quartercaselong} for PPO and ES. We could not reproduce the best results found in Section \ref{sec:ppoversuslegacyextended} within a reasonable amount of time, where 10 experiments were run, and some solutions were better than others. Typically, with the 50,000 samples for ES, we obtained an objective of -9.37 against -6.049 for the best ever, and for PPO, -9.949 against -6.321 for the best ever. This indicates that this problem likely presents very hard-to-reach local optima and emphasizes the need to use statistical techniques to rigorously demonstrate the benefit of introducing a method, as we did in Section \ref{sec:ppoversuslegacyextended} and in our seminal work \cite{seurin2024assessment}.

What matters, however, is that the trends observed in this section are similar to those in the last section, across the other 4 scenarios. Therefore, we can assume that the previous analysis will still hold, and on average, ES and PPO are still better than PESA, and TS is indeed superior to PSA again.

\begin{table}[H]
    \centering
    \begin{tabular}{c|c|c|c|c|c|c|c}
    \hline
 Cases & Obj & Fq & Fh & Cb & Max Bu & Cy & LCOE \\
 \hline
ES &  \textcolor{red}{-6.049} & \textcolor{red}{1.802}  &  \textcolor{red}{1.444}&  \textcolor{red}{\underline{\textbf{1204.2}}} & \textcolor{red}{\underline{\textbf{62.015}}}&\textcolor{red}{\underline{\textbf{497.8}}}&\textcolor{red}{5.257} \\
TS &   -21.628&  \underline{\textbf{1.865}}  &  \underline{\textbf{1.483}}  &  \underline{\textbf{1203.4}} & 59.603  &  \underline{\textbf{496.1}}    &5.314 \\
PSA & -24.669 & \underline{\textbf{1.854}} & \underline{\textbf{1.473}}& \underline{\textbf{1207.9}} & 59.255 & \underline{\textbf{489.1}} & 5.297 \\
PESA & -6.540 &	1.815	& \underline{\textbf{1.456}} &	1153.1	& 60.567 &	\underline{\textbf{497.1}} & 5.269 \\
PESA & -13.64	& 	1.836	& \underline{\textbf{1.468}}	&1169.2	& 58.678& \underline{\textbf{493.3}} & 5.298 \\
PPO& -6.321 & 1.808 & \underline{\textbf{1.455}} & 1174.7 & 60.539 & \underline{\textbf{497.3}} & 5.295 \\
    \end{tabular}
    \caption{Best solution found in the 81-quarter scenario. Highlighted in red is the best case (ES). The violated constraints are underlined and  highlighted in bold.}
     \label{tab:bestfoms81quartercaselong}
\end{table}



\newif\ifusebibtex
\usebibtextrue

\ifusebibtex
\setlength{\baselineskip}{12pt}
\bibliographystyle{mc2023}
\bibliography{mc2023}
\else
\setlength{\baselineskip}{12pt}

\fi

\end{document}